\documentclass[acmsmall]{acmart}

\AtBeginDocument{%
  }

\setcopyright{none}             

\renewcommand\footnotetextcopyrightpermission[1]{}

\acmYear{2026}
\acmMonth{1}
\acmVolume{1}
\acmNumber{1}
\acmArticle{1}

\usepackage{graphicx}%
\usepackage{multirow}%

\usepackage{amsmath,amssymb,amsfonts}
\usepackage{amsthm}%
\usepackage{mathrsfs}%
\usepackage[title]{appendix}%
\usepackage{xcolor}%
\usepackage{textcomp}%
\usepackage{manyfoot}%
\usepackage{booktabs}%
\usepackage{algorithm}%
\usepackage{algorithmicx}%
\usepackage{algpseudocode}%
\usepackage{listings}%
\usepackage{xspace}
\usepackage[acronym]{glossaries}
\newacronym{ml}{ML}{Machine Learning}
\newacronym{ai}{AI}{Artificial Intelligence}
\newacronym{gdpr}{GDPR}{General Data Protection Regulation}
\newacronym{euaiact}{EU AI Act}{European Union's Artificial Intelligence Act}
\newacronym{dnn}{DNN}{Deep Neural Network}
\newacronym{ce}{CE}{Cross-Entropy}
\newacronym{sgd}{SGD}{Stochastic Gradient Descent}
\newacronym{adagrad}{AdaGrad}{Adaptive Gradient Algorithm}
\newacronym{adam}{Adam}{Adaptive Moment Estimation}
\newacronym{rmsprop}{RMSProp}{Root Mean Square Propagation}
\newacronym{ann}{ANN}{Artificial Neural Network}
\newacronym{hpo}{HPO}{Hyperparameters Optimization}
\newacronym{tpe}{TPE}{Tree-structured Parzen Estimator}
\newacronym{nag}{NAG}{Nesterov’s Accelerated Gradient}
\newacronym{fis}{FIS}{Fuzzy Inference System}
\newacronym{gpgpu}{GPGPU}{General-Purpose Graphics Processing Unit}
\newacronym{roc}{ROC}{Reject Option Classification}
\newacronym{moo}{MOO}{Multi-Objective Optimization}
\newacronym{eu}{EU}{European Union}
\newacronym{mse}{MSE}{Mean Squared Error}
\newacronym{mopso}{MOPSO}{Multi-Objective Particle Swarm Optimization}
\newacronym{aco}{ACO}{Ant Colony Optimization}
\newacronym{ga}{GA}{Genetic Algorithm}
\newacronym{pf}{PF}{Pareto Front}
\newacronym{pso}{PSO}{Particle Swarm Optimization}
\newacronym{nsga}{NSGA}{Non-dominated Sorting Genetic Algorithm}
\newacronym{scm}{SCM}{Structural Causal Model}
\newacronym{dag}{DAG}{Directed Acyclic Graph}
\newacronym{ppr}{PPR}{Positive Prediction Rate}
\newacronym{npr}{NPR}{Negative Prediction Rate}
\newacronym{tpr}{TPR}{True Positive Rate}
\newacronym{fpr}{FPR}{False Positive Rate}
\newacronym{tnr}{TNR}{True Negative Rate}
\newacronym{fnr}{FNR}{False Negative Rate}
\newacronym{dp}{DP}{Demographic Parity}
\newacronym{cdp}{CDP}{Conditional Demographic Parity}
\newacronym{ppv}{PPV}{Positive Predictive Value}
\newacronym{ste}{STE}{Straight-Through Estimator}
\newacronym{lse}{LSE}{LogSumExp}
\newacronym{dl}{DL}{Deep Learning}
\newacronym{rf}{RF}{Random Forest}
\newacronym{cv}{CV}{Cross-Validation}
\newacronym{smote}{SMOTE}{Synthetic Minority Over-sampling Technique}
\newacronym{smotenc}{SMOTE-NC}{SMOTE for Nominal and Continuous features}
\newacronym{vm}{VM}{Virtual Machine}
\newacronym{svm}{SVM}{Support Vector Machine}
\newacronym{dt}{DT}{Decision Tree}
\newacronym{hpc}{HPC}{High-performance Computing}
\newacronym{nas}{NAS}{Network Attached Storage}
\newacronym{lvm}{LVM}{Logical Volume Manager}
\newacronym{nmi}{NMI}{Normalized Mutual Information}
\newacronym{gpu}{GPU}{Graphics Processing Unit}
\newacronym{cnn}{CNN}{Convolutional Neural Network}
\newacronym{fadol}{FADOL}{Fair and Accurate Dual Optimization Loss}
\newacronym{compas}{COMPAS}{Correctional Offender Management Profiling for Alternative Sanctions}
\newacronym{cs}{Computer System}{CS}
\newacronym{eo}{Equalized Odds}{EO}

\glsdisablehyper
\usepackage{longtable} 
\usepackage{booktabs}
\usepackage{array}     
\usepackage{xltabular} 
\usepackage{enumitem}
\usepackage{multicol}

\usepackage{setspace}
\usepackage{calligra}
\usepackage[T1]{fontenc}
\usepackage{etoolbox}
\usepackage{microtype}
\usepackage{booktabs}   
\usepackage{multirow}   
\usepackage{colortbl}   
\usepackage{xcolor}     
\usepackage{adjustbox}  
\usepackage{lscape}     
\makeglossaries

\newcommand{\acro}{\textsc{FairMetrics}\xspace}

\usepackage{balance}

\usepackage[normalem]{ulem}

\theoremstyle{plain}%
\newtheorem{theorem}{Theorem}%
\newtheorem{proposition}[theorem]{Proposition}

\newtheorem{example}{Example}%
\usepackage{amsthm}
\usepackage{cleveref}

\theoremstyle{definition} 
\newcommand{\proofstep}[1]{%
    \par 
    \noindent\textit{#1.}\quad
}
\theoremstyle{thmstylethree}%

\newcommand{\independent}{\mathrel{\perp\!\!\!\perp}}

\theoremstyle{remark} 
\newtheorem{numberedproof}{Proof}

\usepackage{cleveref}
\crefname{numberedproof}{Proof}{Proofs}
\Crefname{numberedproof}{Proof}{Proofs}

\usepackage{xcolor}
\usepackage{colortbl}
\usepackage{collcell}
\usepackage{pgf}
\usepackage{xstring}
\newcommand{\colorgap}[1]{%
  \StrDel{#1}{$}[\cleanval]%
  \pgfmathsetmacro{\val}{\cleanval}%
  \pgfmathparse{\val < 0 ? 1 : 0}%
  \ifdim \pgfmathresult pt = 1 pt %
    \pgfmathsetmacro{\pct}{min(60, max(0, (-60 * \val) / 0.60))}%
    \edef\temp{\noexpand\cellcolor{red!\pct!white}}%
    \temp #1%
  \else
    \pgfmathsetmacro{\pct}{min(60, max(0, (60 * \val) / 0.15))}%
    \edef\temp{\noexpand\cellcolor{blue!\pct!white}}%
    \temp #1%
  \fi
}
\newcolumntype{G}{>{\collectcell\colorgap}c<{\endcollectcell}}

\begin{document}

\title{Rethinking Fairness Metrics for Multi-Class and Multi-Group Assessment: Not Everything is Binary}

\author{Matteo Grazioso}
\correspondingauthor
\authornotemark[1]
\email{matteo.grazioso@unive.it}
\orcid{0009-0004-2802-7501}
\affiliation{%
  \institution{Department of Environmental Sciences, Informatics and Statistics, Ca' Foscari University of Venice}
  \city{Venice}
  \country{Italy}
}

\author{Fabio Naccarella}
\authornote{Both authors contributed equally to this research.}
\email{900619@stud.unive.it}
\orcid{0009-0003-5313-634X}
\affiliation{%
  \institution{Department of Environmental Sciences, Informatics and Statistics, Ca' Foscari University of Venice}
  \city{Venice}
  \country{Italy}
}

\author{Chiara Gallese}
\email{C.Gallese@tilburguniversity.edu}
\orcid{0000-0001-8194-0261}
\affiliation{%
  \institution{Tilburg Institute for Law, Technology, and Society, Tilburg University}
  \city{Tilburg}
  \country{The Netherlands}
}

\author{Marco S. Nobile}
\correspondingauthor
\email{marco.nobile@unive.it}
\orcid{0000-0002-7692-7203}
\affiliation{%
  \institution{Department of Environmental Sciences, Informatics and Statistics, Ca' Foscari University of Venice}
  \city{Venice}
  \country{Italy}
}
\affiliation{%
  \institution{European Center for Living Technologies (ECLT)}
  \city{Venice}
  \country{Italy}
}

\renewcommand{\shortauthors}{Grazioso et al.}

\begin{abstract}
Artificial Intelligence, and Machine Learning in general, is becoming pervasive across all domains and disciplines.
By means of this data-driven approach, algorithms can learns complex task just by learning from samples.
However, if the data-set is biased, they might lead to the development of unfair computer systems, that is, characterised by a prejudice towards specific individuals or groups, which might get an unmotivated output from the system.
This circumstance lead to the development of multiple metrics designed to quantify the level of unfairness of an algorithm.
However, a generalised assumption in the literature is that the sensitive features and the output labels are binary.
When the data is not binary, it is forcefully binarized using a one-versus-the-rest simplification.
In this work, we discuss the limitations of existing metrics and propose novel, mathematically sound, and generalized multi-group and multi-class formulations of the metrics. Ultimately, our results highlight the empirical assessment distortion caused by binarization strategies, which reveals the limitations of binary fairness metrics in the multi class setting.
\end{abstract}




\received{20 February 2007}
\received[revised]{12 March 2009}
\received[accepted]{5 June 2009}

\maketitle

\section{Introduction}
\label{sec:intro}
Computer Systems have long been known to embed values and reproduce existing biases, leading to discriminatory outcomes, due both to the way they are developed and the way they are used. 
As Friedman and Nissenbaum argued \cite{friedman1996bias}, they may systematically disadvantage individuals or groups through three different biases: pre-existing bias, rooted in social institutions and practices; technical bias, arising from design constraints and formalisation choices; emergent bias, which materialises only in the context of use. 
Models trained on historical records may inherit the patterns of disadvantage sedimented in them, even in absence of discriminatory intent or explicit reliance on protected characteristics \cite{barocas2023fairness,wachter2021fairness}. 
Algorithmic decision-making thus brings epistemic and normative issues, such as unfair outcomes, that cannot be dissociated from its social context \cite{mittelstadt2016ethics}.

Notwithstanding this, in recent years the adoption of data-driven systems for decision making has skyrocketed, consistently spreading across all disciplines; advanced \gls{ai} and \gls{ml} models were introduced into high-stakes domains, sometimes achieving remarkable performances \cite{carriero2025ai}, some other times bringing negative consequences \cite{engels2025code}.
For example, \gls{ai} systems have been used to screen \textit{Curriculum Vitae} of job candidates \cite{al2021role}, assess cardiac functionality \cite{nobile2025assessing}, identify suspects in criminal investigations \cite{gallese2026predictive}, or for border control purposes \cite{bor2023balancing}.
\gls{ai} systems can even be deployed in autonomous weapons \cite{falletti2022ethical}.
Such a large-scale adoption raised significant concerns, for example regarding data protection, trust, reliability, or transparency.

To address these concerns, some modern legislative frameworks enforce strict provisions and requirements for developers, deployers and users. 
One example is the \gls{gdpr}~\cite{EU2016_GDPR}. 
Article 22 restricts decisions based solely on automated processing that produce legal or similarly significant effects. Article 15(1)(h) grants access to meaningful information about the logic involved. 
The reference to an explanation of the decision appears only in non-binding Recital 71. Nevertheless, the Court of Justice has held that Article 15(1)(h) requires an intelligible explanation of the procedure and principles actually applied, sufficient to enable the data subject to contest the decision (C-203/22, Dun and Bradstreet Austria, 2025).
The \gls{euaiact}~\cite{act2024eu} identifies a number of \gls{ai} fields that impact citizens' fundamental rights and liberties, classifying them as high-risk systems, and imposing stringent requirements, such as data-governance duty to examine data for possible biases and to take appropriate measures, interpretability of the system's output, appropriate usage of their models, traceability and  monitoring of the systems throughout their lifecycle.
According to Regulation (EU) 2026/1744 (Digital Omnibus on AI), these obligations apply to Annex III systems from 2 December 2027. 
However, neither instrument specifies how bias is to be measured, and the choice of fairness criteria and metrics is left to providers, to standardisation and, ultimately, to the interpretation of Union non-discrimination law.

These obligations, however, do not by themselves ensure that automated decisions are non-discriminatory. Algorithmic discrimination typically operates through proxy variables rather than through explicit reliance on protected characteristics. It is therefore predominantly the doctrine of indirect discrimination that is engaged. Under Union law, indirect discrimination occurs where an apparently neutral provision, criterion or practice would put persons sharing a protected characteristic at a particular disadvantage compared with other persons, unless it is objectively justified by a legitimate aim and the means of achieving that aim are appropriate and necessary (Article 2(2)(b) of Directive 2000/43/EC; Article 2(2)(b) of Directive 2000/78/EC; Article 2(1)(b) of Directive 2006/54/EC). Since the definition refers to a disadvantage that a practice would produce, it does not presuppose that harm has materialised. It can therefore be assessed ex ante, which is precisely the function of a fairness audit.

From a technical perspective, within discriminatory issues that arise from \glspl{cs} development, research has identified problems related to training data and model or feature selection. 
To identify, mitigate, and overcome these problems, a variety of fairness checks have been proposed both at the data and the model level.
However, fairness is a concept that is difficult to achieve and whose definition is not agreed upon. Research has shown that different fairness concepts can be incompatible with each other and the most prominent metrics cannot in general be jointly satisfied, making perfect fairness a difficult target \cite{kleinberg2016inherent,chouldechova2017fair}. A further problem concerns the dimensionality of the decision itself. The debate has been shaped by binary paradigms, but many legally relevant decisions are not binary (credit tiers, triage levels, academic outcomes), and protected characteristics are frequently multi-valued and intersecting \cite{crenshaw1995mapping}.

What is required, therefore, is a reasoned selection of the criterion appropriate to the decisional context, together with a measurement instrument that faithfully represents the disparities relevant to it. Barocas, Hardt and Narayanan \cite{barocas2023fairness} trace group-fairness metrics to three observational criteria: Independence, Separation and Sufficiency. 
Building on their account, we adopt what we term a criteria-first approach, in which metrics are derived from, and evaluated against, the criterion they operationalise. Specifically, we extend binary group fairness metrics to multi-class and multi-group settings under Barocas \textit{et al.}'s foundational probabilistic criteria; extend theoretical proofs regarding the mutual exclusivity of these criteria to multi-class scenarios; empirically quantify the distortion gaps induces by binarization paradigms and finally introduce an open-source library for multi-class and multi-group fairness auditing.
The framework does not propose to automate the legal assessment of discrimination. It seeks, rather, to ensure that the statistical evidence on which such an assessment may rely is computed on the decision as actually taken.

The remainder of this manuscript is structured as follow.
Section~\ref{sec:relWorks} review the State-of-the-Art on algorithmic fairness and existing classification metrics.
Section~\ref{sec:LABELSEC3} presents our multi-class metrics extensions categorized by foundational probabilistic criteria.
Section~\ref{sec:mutal_exclusion} discusses the mutual exclusivity and co-existence of the aforementioned criteria.
Section~\ref{sec:experimental} first introduces our open-source library for binary, multi-class, and multi-group fairness evaluation, followed by the experimental setup and empirical comparison between native multi-class assessment and binary reduction approaches.
Finally, Section~\ref{sec:conclusion} draws conclusions and discusses future research directions.

\section{Related Works}
\label{sec:relWorks}
The scientific literature lacks consensus on the terms fairness, bias and discrimination, terms that are often used interchangeably \cite{mehrabininareh10galstyan,gallese2026predictive,gallese2025fanfair}. 
However, the metrics presented in this paper assume a definite relation among these terms, so we will disambiguate  their meaning as follows.

\textit{Bias} -- Following Friedman and Nissenbaum \cite{friedman1996bias}, bias denotes a systematic and unfair tendency of a computer system to disadvantage certain individuals or groups in favour of others. It is distinct from statistical bias, \textit{i.e.}, the systematic deviation of an estimator from the quantity it estimates. 
Bias may originate at any stage of the ML lifecycle \cite{suresh2021framework}. 
Because sensitive information may be redundantly encoded in other features (e.g., proxy variables), excluding sensitive attributes may not remove bias \cite{barocas2023fairness,wachter2021fairness}. 
Moreover, automated systems may operate within feedback loops in which biased outputs shape future training data ~\cite{barocas2016big,richardson2019dirty}.

\textit{Discrimination} -- It is a legal concept that varies depending on the specific legal system. European Union law recognises two principal forms: direct discrimination occurs where a person is treated less favourably than another is, has been or would be treated in a comparable situation, on a protected ground. It admits only the exceptions expressly provided by law;
indirect discrimination occurs where an apparently neutral provision, criterion or practice puts persons sharing a protected characteristic at a particular disadvantage compared with others, absent objective justification \cite{wachter2021fairness}.
Bias is thus a possible cause of discrimination, but it is not an equivalent concept.
A statistical disparity, that is, the probability of an outcome is different across demographic groups, does not necessarily constitute discrimination. 
It may, however, establish facts from which discrimination can be presumed, shifting to the respondent the burden of proving that the principle of equal treatment has not been breached (Article 8 of Directive 2000/43/EC). 
Whether the disadvantage is ``particular'' and whether it is justified remain contextual determinations is ultimately reserved to the courts \cite{wachter2021fairness}.

\textit{Fairness} -- This is the most contested of the three notions.
In the technical literature, it is commonly defined as a quantity to be minimized, as the absence of prejudice or favouritism towards individuals or groups on the basis of inherent or acquired characteristics \cite{mehrabininareh10galstyan}. 
Operationally, fairness is rendered through formal criteria that constrain the joint distribution of predictions, outcomes and group membership \cite{mittelstadt2016ethics}. 
Each criterion encodes a specific conception of equality, and the choice ultimately depends on the value systems of a society \cite{wachter2022theory,wachter2021bias,wachter2021fairness}.

Two further notions recur throughout this paper. 
First, we call \emph{sensitive} attribute a variable that encodes the membership to groups defined by a protected ground. 
\textit{Protected grounds} are legal categories in the domain of anti-discrimination law (see, \textit{e.g.}, Article 19 TFEU and Article 21 of the Charter of Fundamental Rights). 
A sensitive attribute is the representation of such categories in data, which may be absent, inferred or proxied. 
Although the terms sensitive and protected are often used interchangeably, they should be kept distinct.
Neither notion coincides with the special categories of personal data under Article 9 GDPR, which include racial or ethnic origin and health but not sex or age.
Second, control variables (or strata) are features regarded as legitimate grounds for differentiated outcomes, such as qualifications or credit history. Conditioning on them yields conditional variants of the fairness criteria \cite{corbett2017algorithmic}.
Their legitimacy is itself a normative determination, functionally analogous to objective justification in indirect discrimination. Moreover, conditioning on a variable that acts as a proxy for the protected ground may mask disparity \cite{wachter2021fairness,wachter2021bias}.

Mitigation techniques are conventionally classified as pre-, in- or post-processing, according to whether they intervene on the data, on the learning objective, or on the predictions of a trained model. 
Each approach implies the measurement of a  disparity to be corrected. 
Data-level instruments such as FanFAIR \cite{gallese2025fanfair,gallese2025dataset,rispoli2025investigating} complement model-level metrics in this respect. 
For a more detailed analysis, prior literature provides foundational overviews~\cite{sekwenz2024measuring,castelnovo2022clarification,wachter2021fairness,mehrabininareh10galstyan}.

Scientific literature divides algorithmic fairness into three primary paradigms: individual fairness, group fairness, and subgroup fairness~\cite{castelnovo2022clarification, mehrabininareh10galstyan}. 

\textit{Individual fairness} formalised by Dwork \textit{et al.} \cite{dwork2012fairness}, requires that similar individuals be treated similarly. 
The requirement restates the precept that like cases be treated alike, which Hart identified as the core of the idea of justice \cite{hartlaw1961concept}.
Hart also observed that the precept remains empty until the criteria of relevant resemblance and difference are fixed \cite{hartlaw1961concept}.
Individual fairness inherits this incompleteness.
Dwork \textit{et al.} acknowledge that the similarity metric is the crux of the approach and must be supplied externally, and specifying it requires a normative consensus rarely available in practice \cite{dwork2012fairness}.

\textit{Group fairness} partitions the population according to the sensitive attribute and requires equality of selected statistics across groups. 
It is the dominant paradigm \cite{rabonato2025systematic}, which is partly due its computational tractability, and partly due to its structural analogy with the comparative reasoning of indirect discrimination, 
an analogy that should not, however, be mistaken for equivalence. 
Statistical parity neither establishes nor excludes discrimination, and mandates a further assessment of particular disadvantage \cite{wachter2021fairness}.

\textit{Subgroup fairness} responds to what Kearns \textit{et al.} call fairness gerrymandering, by analogy with the electoral practice of drawing district boundaries to secure a predetermined outcome. A classifier may satisfy a parity constraint for each protected attribute taken separately, yet violate it severely on subgroups defined by combinations of attributes, such as Black women~\cite{kearns2018preventing}. Kearns \textit{et al.} therefore require parity, for statistical parity and for false positive rates, not over a few predefined groups but over every subgroup in a class $\mathcal{G}$ defined by functions of the protected attributes. An example of such a class is all conjunctions of attribute values, which can contain exponentially many subgroups. Each subgroup’s disparity is weighted by its size. \cite{kearns2018preventing}. 
In addition, the number of intersectional subgroups grows combinatorially with the number of attributes. Many subgroups therefore contain too few individuals for their rates to be estimated reliably, which is why Kearns \textit{et al.} weight disparities by subgroup size \cite{kearns2018preventing}.
The present work adopts an intermediate position: intersectional groups are represented explicitly as the Cartesian product of sensitive attributes, within the group-fairness paradigm.

Concerning \emph{group} fairness, scholars generally rely on formalizations based on probability theory. 
In particular, seminal literature establishes three criteria for classification problems based on the conditional independence relations among three random variables, namely the sensitive attribute, the true target label, and the predicted label~\cite{barocas2023fairness}.
These criteria are: \textit{Independence}, \textit{Separation}, and \textit{Sufficiency}, which are defined in details in Section~\ref{subsec:rebuild_Independence},~\ref{subsec:rebuild_Separation} and~\ref{subsec:rebuild_Sufficiency}, respectively.
To simplify the understanding of such criteria, we provide Examples~\ref{example:loan_approval1}, \ref{example:loan_approval2}, \ref{example:loan_approval3}, \ref{example:loan_approval4}, and \ref{example:loan_approval5} in Appendix~\ref{app:secA}.
We stress the fact that not all criteria are desirable in most scenarios.
As an example, clinicians do not require an unjustified equal illness prediction rate across genders, that is, the Independence, because the prevalence can be different.
On the contrary, both Separation and Sufficiency are typically more relevant in clinical applications.
It is worth noting that criteria can be mutually exclusive in real world scenarios. Barocas \textit{et al.} formally proved that: Independence and Sufficiency are incompatible \cite{barocas2023fairness}; Separation and Sufficiency are also incompatible, under the additional assumption that all events have positive probability; Independence and Separation are incompatible in the binary case.
Hence it is crucial for practitioners to select the correct metrics, which is problem-dependent.

Kleinberg \textit{et al.} and Chouldechova obtained analogous results for calibration and error-rate balance \cite{kleinberg2016inherent,chouldechova2017fair}. The famous COMPAS risk assessment tool is a prominent example: according to a ProPublica investigation \cite{larson2016we}, Black defendants who did not reoffend were classified as higher risk at nearly twice the rate of white defendants. The developer replied that the scores were equally predictive across groups \cite{angwin2022machine,corbett2017algorithmic}. 
Both claims can be true because, under unequal base rates, the error-rate parity and the predictive parity diverge \cite{chouldechova2017fair}. Since the precision of a prediction depends on the prevalence of the outcome in the group to which it is applied, a classifier with equal error rates across groups will, when base rates differ, generally yield unequal predictive values, and vice versa \cite{chouldechova2017fair}.
The selection of a criterion is therefore an unavoidable, problem-dependent normative choice. 
Interestingly, we show in Section~\ref{sec:mutal_exclusion} that, in the multi-class setting, Independence and Separation can coexist under very specific conditions.

These  probabilistic criteria translate directly into the vast ``zoo'' of fairness metrics \cite{castelnovo2022clarification}.
Historically, researchers focused almost exclusively on binary classification problems, because the simple two-class setting drastically minimizes the complexity of the underlying probability distributions.
As a matter of fact, multi-class and multi-group settings have received considerably less attention \cite{alghamdi2022beyond} and current literature proposes few extensions such as Demographic Parity~\cite{purificato2024toward,denis2024fairness}, Equalized Odds~\cite{hardt2016equality},  multi-class Equal Opportunity~\cite{purificato2024toward, rouzot2022learning} or Overall Misclassification~\cite{rouzot2022learning}.
Section~\ref{sec:LABELSEC3} provides a detailed analysis of prior literature and our novelty, whereas Table~\ref{tab:fairness_metrics} summarizes the core of this manuscript: presenting the well established binary metric landscape, outlining our proposed multi-class and multi-group taxonomy, and highlighting our contributions to the current literature.

However, modern applications need to shift towards multi-class and multi-group classification landscape.
Multi-category scenarios are pretty frequent in real-world applications, but are treated less thoroughly~\cite{alghamdi2022beyond}, since the extension to such cases introduce  mathematical challenges and can spark  semantic contradictions due to the use of misleading or conflicting names for the same metrics.
This work emerges from that conflict and addresses these structural gaps, as shown in 
Table~\ref{tab:fairness_metrics}.
The table summarises the state-of-the-art metrics, categorizing the existing multi-class metrics, organized under the three foundational probabilistic criteria.
Furthermore, the table shows the major gaps in current literature, particularly in the multi-class scenario.
To resolve these limitations, our framework proposes novel metrics that complete the taxonomy, while Section~\ref{sec:LABELSEC3} provides the mathematical definitions and formal implementations for each metric listed in the overview.

\begin{table}[!htbp]
\centering

\begin{tabularx}{\textwidth}{>{\bfseries\raggedright\arraybackslash}l >{\raggedright\arraybackslash}X >{\raggedright\arraybackslash}X >{\raggedright\arraybackslash}X}
\toprule
\textbf{Criterion} & \textbf{Established Binary Classification Taxonomy} & \textbf{Proposed Multi-class \& Multi-group Classification Taxonomy} & \textbf{Earlier Multi-class and Multi-group Works and Our Contributions} \\
\midrule

\multirow{3}{*}{Independence}
& Demographic Parity~\cite{mehrabininareh10galstyan,pessach2022review,rabonato2025systematic,castelnovo2022clarification}
& Demographic Parity
& Proposed in~\cite{purificato2024toward,denis2024fairness} \\
& Conditional Demographic Parity~\cite{mehrabininareh10galstyan,castelnovo2022clarification}
& \cellcolor{gray!15}Conditional Demographic Parity
& \cellcolor{gray!15}\textbf{Proposed in this Work} \\
\midrule

\multirow{8}{*}{Separation}
& Equalized Odds~\cite{mehrabininareh10galstyan,pessach2022review,rabonato2025systematic,castelnovo2022clarification}
& Equalized Odds
& Proposed in~\cite{hardt2016equality} \\
& Equal Opportunity~\cite{mehrabininareh10galstyan,pessach2022review,rabonato2025systematic,castelnovo2022clarification}
& \textbf{\textit{Not generalizable}}
& \\
& Predictive Equality~\cite{castelnovo2022clarification}
& \textbf{\textit{Not generalizable}}
& \\
& \textbf{\textit{Not Binarizable}}
& Diagonal Equality
& Proposed in~\cite{purificato2024toward,rouzot2022learning} \\
& \textbf{\textit{Not Binarizable}}
& \cellcolor{gray!15}Off-diagonal Equality
& \cellcolor{gray!15}Partially Proposed in~\cite{rouzot2022learning}; \textbf{Complete Extension Proposed in this Work} \\
\midrule
Sufficiency
& Predictive Parity~\cite{castelnovo2022clarification}
& \cellcolor{gray!15}Multi-class Predictive Parity
& \cellcolor{gray!15}\textbf{Proposed in this Work} \\
\midrule
Accuracy
& Equal Accuracy~\cite{castelnovo2022clarification}
& \cellcolor{gray!15}Equal Accuracy
& \cellcolor{gray!15}\textbf{Proposed in this Work} \\
\bottomrule

\end{tabularx}
\caption{Current and proposed taxonomy of fairness metrics categorized by foundational probabilistic criteria for binary, multi-class, and multi-group classification settings. Shaded cells highlight our contributions.}
\label{tab:fairness_metrics}
\end{table}

To overcome the current insufficiency of multi-class metrics, practitioners can deploy a one-vs-rest target binarization strategy to reduce a multi-class classification task into a sequence of binary cases, which can be evaluated using established binary metrics.
Although this decomposition simplifies computational execution, it creates a significant blind spot because it merges all remaining categories into a single, heterogeneous baseline group which potentially hides the unique disparities and distinct error profiles that exist between the remaining individual classes, as discussed in Section~\ref{sec:experimental}.

\section{From Binary to Multi-Class and Multi-Group Fairness Assessment}
\label{sec:LABELSEC3}
This Section peruses two primary objectives: \textit{(i)} mapping empirical evaluation metric to to their fairness criterion, hence reorganizing them in a structured unified taxonomy, and \textit{(ii)} generalizing binary formulations to multi-group and multi-class settings;
We first introduce the notation and then detail the fairness criteria and the corresponding metrics; Appendix~\ref{app:secA} provides illustrative scenarios for each metrics (Examples~\ref{example:loan_approval2},
\ref{example:loan_approval3},
\ref{example:loan_approval4},
\ref{example:loan_approval5},
\ref{example:eq_odds_vs_eq_acc_repayingLoan},
\ref{example_equal_opp_disease_detection},
\ref{example:pred_equality_criminal},
\ref{example:diag_eq_medical_triage},
\ref{example:off-diag-eq}, \ref{example:bin_pred_parity}, \ref{example:mul_pred_parity}, and  \ref{example:equal_acc}).

\subsection{Notation and Multi-Class Extension Design Choices}
\label{subsec:notation}
\subsubsection{Notation}
Let $(\Omega, \mathcal{F}, \mathbb{P})$ be a probability space, where $\Omega$ represents the sample space of all possible individuals (or instances), $\mathcal{F}$ is the set of all possible events, and $\mathbb{P}$ is the probability measure. 
Over this space, we define a set of random variables that characterize our classification problem.
Let $X: \Omega \rightarrow \mathcal{X} \subseteq \mathbb{R}^d$ be a random vector representing the observable, non-sensitive features of an individual, namely the Input Feature Space, where $d \in \mathbb{N}^+$, denotes the total number of input features.
Let $A: \Omega \rightarrow \mathcal{A}$ be a discrete random variable representing the protected or sensitive features of an individual (\textit{e.g.}, gender or ethnicity).
Our framework explicitly supports multi-valued attributes and intersectional groups, such that $\mathcal{A} = \{a_1, a_2, \dots, a_m\}$ where $|\mathcal{A}| = m \ge 2$.
For conditional fairness criteria, we define the Control Variable Space $R: \Omega \rightarrow \mathcal{R}$ which represents a set of legitimate control variables or ``strata''.
Similarly, our framework supports multi-valued and intersectional Control Variables, such that $\mathcal{R} = \{r_1, r_2, \dots, r_m\}$ where $|\mathcal{R}| = L \ge 2$.
Let $Y: \Omega \rightarrow \mathcal{Y}$ denote the target variable that the model aims to predict.
Our framework support arbitrary label space cardinality, such that the set of categorical labels $\mathcal{Y} = \{1, 2, \dots, K\}$, where $K \ge 2$.
For the binary classification task, we define $\mathcal{Y} = \{0, 1\}$.
We define a classifier as a function $f$, that maps input features $(\mathcal{X},\mathcal{A})$ to a probability distribution over the $K$ classes; the discrete prediction generated by the model is represented by the random variable $\hat{Y} \in \mathcal{Y}$.
The supervised learning and fairness evaluation process operates over a dataset $\mathcal{D} = \{(x_i, a_i, r_i, y_i, \hat{y}_i)\}_{i=1}^N$, assumed to be sampled independently and identically distributed (\textit{i.i.d.}) from the joint probability distribution $\mathbb{P}(X, A, R, Y, \hat{Y})$.

\subsubsection{Extension Design Choices}
Most foundational fairness metrics were originally designed for binary classification settings, involving only binary sensitive attributes, \textit{i.e.,} between two demographic groups (\textit{e.g.} male and female cohorts).
Transitioning to the case of multi-class / multi-group and  classification requires some strategies to generalize binary metrics without obscuring class-specific or group-specific bias.

\paragraph{Single Sensitive Feature Logic and Intersectionality}
Real-world audits often require evaluating individuals at the intersection of multiple sensitive attributes.
Given a vector of attributes $A_{vec} = [A_1, A_2, \dots, A_n]$, we compute the Cartesian product of their possible assignments to construct a unified categorical space $\mathcal{A}_{comb}$. 
That is, we map an $(N_{\text{entities}} \times N_{\text{features}})$ attribute matrix into a  vector where each unique intersectional profile is mapped to a single distinct integer.

\paragraph{Multi-Dimensional Disparity}
When the outcome space is artificially simplified via naive binarization or one-vs-rest reduction approaches, fairness constraints on a single target fail to account for how misclassifications are partitioned across the remaining categories.
We can refer to this blind spot as \textit{Multi-Dimensional Disparity}.
To address Multi-Dimensional Disparity, we adopt a Max-Diff Aggregation approach~\cite{putzel2022blackbox}.
This formulation is deeply rooted in the Rawlsian principle of justice, which states that the ethical acceptability of a system is bounded by its treatment of the most disadvantaged demographic group~\cite{edor2020john}.
Therefore, the specific metric computes the disparity for every class and returns the maximum observed difference across any pair of groups, thus identifying the worst-case disparity.
Specifically, we follow the common approach in the fairness literature and measure the largest absolute disparity across all protected groups.
Formally, for a given fairness metric, we define


\begin{equation}
\max_{k \in \mathbf{\mathcal{\hat{Y}}}} \max_{j \in \mathbf{\mathcal{Y}}} \max_{a,b \in \mathcal{A}} \max_{r \in \mathbf{\mathcal{R}}} \mathbf{\lvert} P_{k,j,a,r}-P_{k,j,b,r} \mathbf{\rvert},
\label{eq:base_fairness_metric}
\end{equation}
where $P_{k,j,a,r}$ denotes the probability associated with predicted class $k$ and true class $j$ for protected group $a$, whose definition depends on the fairness metric under consideration; $r$, instead, is the control variable strata, present only in the case of Conditional Demographic Parity.
Since every probability satisfies $P_{k,j,a,r}\in[0,1]$, each fairness metric considered in this work also takes values in the interval $[0,1]$, where lower values indicate smaller disparities across groups and therefore greater fairness.

Although it is mathematically possible to achieve null fairness disparities, in real-world scenarios it is very unlikely.
Consequently, practitioners commonly select a threshold value $\varepsilon$ which identifies the maximum fairness disparity tolerable, \textit{i.e.}, \gls{ml} models which score $\text{Fairness} < \varepsilon$ are considered fair enough.

Although in this work we only focus on fairness metrics based on fairness discrepancy between distinct groups, other strategies exist (e.g., Treatment Equality~\cite{berk2021fairness}, Disparate Impact~\cite{feldman2015certifying})
Alghamdi \textit{et al.}, proposed multi-class extensions for Statistical Parity, Equalized Odds, and Accuracy Equality based on the relative deviation of group-conditional probabilities from the overall population average, bounded by a prescribed tolerance parameter $\alpha$~\cite{alghamdi2022beyond}.

\subsection{Rebuilding from the Criteria: Independence}
\label{subsec:rebuild_Independence}
The first group fairness criterion is \textit{Independence}, which requires that the predictive outcome $\hat{Y}$ operates completely independently of the sensitive attribute $A$, \textit{i.e.}
$
    \hat{Y} \independent A.
$
This criterion ignores the ground truth $Y$, which makes it suitable in scenarios where the true label is unavailable or cannot be reliably trusted.
Enforcing \textit{Independence} requires equalizing class selection rates among protected groups, regardless of the ground truth original distribution; consequently, it is said to be a \textit{non-conservative criterion} because it often forces a change in the status quo, \textit{i.e.}, the true target distribution of each group.
Unless the ground truth is perfectly independent of the sensitive attribute, which rarely holds in practice, a ``perfect classifier'' -- a classifier that commits no error -- generally does not satisfy \textit{Independence}.

\subsubsection{Demographic Parity}
Independence can be evaluated by the group-fairness metric defined in the literature as \gls{dp}, also known as \textit{Statistical Parity} or simply \textit{Group Fairness}.
This metric demands equal selection rates across protected groups for each target.

In the case of binary classification \cite{mehrabininareh10galstyan,pessach2022review,rabonato2025systematic,castelnovo2022clarification}, \gls{dp} is defined as:
\begin{equation}
        P(\hat{Y} = 1 \mid A = a) = P(\hat{Y} = 1 \mid A =b) \quad \forall a,b \in \mathcal{A} .
\end{equation}
The binary metric requires equal \gls{ppr}, \textit{i.e.,} favourable outcomes, across every group.
In the binary setting, demanding equal \gls{ppr} across protected groups implies equal  \gls{npr}.
Scholars proposed the natural extension of \gls{dp} to support Independence in multi-class, multi-group classification scenario~\cite{purificato2024toward,denis2024fairness}:
\begin{equation}
    P(\hat{Y} = \hat{y} \mid A = a) = P(\hat{Y} = \hat{y} \mid A =b) \quad \forall \hat{y} \in \mathcal{Y}, \forall a,b \in \mathcal{A} .
\end{equation}
In this formulation, \gls{dp} proves to be the direct implementation of the Independence criterion, requiring equal selection rates across protected groups for each class.
Binary \gls{dp} is clearly a special case of the multi-class \gls{dp} where the labels are limited to $\mathcal{Y} = \{0, 1\}$.

\subsubsection{Conditional Demographic Parity}
\gls{cdp}, also known as \textit{Conditional Statistical Parity}, is an extension of the Independence criterion that requires model's predictions $\hat{Y}$ to be independent of the sensitive attribute $A$, given a specific set of control variables known as ``strata'' $R$:
$
    \hat{Y} \independent A \mid R.
$
Similar to Binary \gls{dp}, binary \gls{cdp} was originally defined by comparing only \gls{ppr}~\cite{mehrabininareh10galstyan,castelnovo2022clarification}:
\begin{equation}
    P(\hat{Y} = 1 \mid A = a, R = r) = P(\hat{Y} = 1 \mid A =b, R = r) \quad \forall a,b \in \mathcal{A}, \forall r \in \mathcal{R}.
\end{equation}
Similarly to \gls{dp}, it implies equal \gls{npr}.

In the existing literature, multi-class \gls{cdp} has received significantly less attention than the other fairness metrics.
For that reason, in this work we propose the following multi-class and multi-group \gls{cdp} generalisation to fill the current gap in the literature:
\begin{equation}
    P(\hat{Y} = \hat{y} \mid A = a, R = r) = P(\hat{Y} = \hat{y} \mid A =b, R = r) \quad \forall \hat{y} \in \mathcal{Y}, \forall a,b \in \mathcal{A}, \forall r \in \mathcal{R}.
\end{equation}
Once again, binary \gls{cdp} is a special case of the multi-class \gls{cdp} where $\mathcal{Y} = \{0, 1\}$.

\subsection{Rebuilding from the Criteria: Separation}
\label{subsec:rebuild_Separation}
The \textit{Separation} criterion requires the prediction $\hat{Y}$ and the sensitive attribute $A$ to be conditionally independent given the true label $Y$, \textit{i.e.}
$
    \hat{Y} \independent A \mid Y.
$
Unlike \textit{Independence}, \textit{Separation} requires information about the target $Y$, which means it could potentially perpetuate pre-existing bias present in the ground truth $Y$.
Moreover, \textit{Separation} is a ``conservative'' criterion, meaning that a ``perfect classifier'' ($\hat{Y} = Y$)  always satisfies the criterion.

\subsubsection{Equalized Odds}
\label{sec:equalized_odds}
\gls{eo} is the group-fairness metrics that evaluates compliance with Separation for a \gls{ml} model's predictions.
\gls{eo} is satisfied when the confusion matrix looks statistically similar for every sensitive group.
Indeed, it is commonly defined as the \textit{error-parity metric}~\cite{castelnovo2022clarification}.
It is worth noting that this requirement is strictly stronger than simply imposing equal overall accuracy across protected groups.
While EO enforces equal rates for each specific error type among groups, equal overall accuracy aggregates them checking for equal unconditioned misclassification rates.
To illustrate this distinction we provide Example~\ref{example:eq_odds_vs_eq_acc_repayingLoan} in Appendix~\ref{app:secA}.
Alternative fairness metrics focusing solely on equal accuracy have been proposed in the literature; they will be described in detail in Section~\ref{subsec:rebuild_Accuracy}.


For a binary outcome, \gls{eo} \cite{mehrabininareh10galstyan,pessach2022review,rabonato2025systematic,castelnovo2022clarification} requires both the \gls{tpr} and \gls{fpr} to be equal across all protected groups, \textit{i.e.}:
\begin{equation}
    P(\hat{Y} = 1 \mid A = a, Y = y) = P(\hat{Y} = 1 \mid A =b, Y = y) \quad \forall a,b \in \mathcal{A}, \forall y \in \{0,1\}.
\end{equation}
The literature provides two relaxations for the binary metrics:
\begin{itemize}
    \item Equal Opportunity, which focuses on \gls{tpr} ($Y=1$) and guarantees equal \gls{fnr} as a byproduct;
    \item Predictive Equality, which focuses on \gls{fpr} ($Y=0$) and guarantees equal \gls{tnr} as a byproduct.
\end{itemize}
Both relaxations will be further analysed in Section \ref{sec:sep_relaxations}.

As observed by Hardt \textit{et. al}~\cite{hardt2016equality}, the general definition of the Separation criterion already account for both binary or multi-class targets and sensitive features.
Consequently, in the multi-class and multi-group scenario, EO is defined as follows:
\begin{equation}
    P(\hat{Y} = k \mid A = a, Y = j) = P(\hat{Y} = k \mid A =b, Y = j) \quad \forall a,b \in \mathcal{A}, \forall j,k \in \mathcal{Y}.
\label{eq:eq_odds}
\end{equation}
Binary \gls{eo} represents a special case of the multi-class formulation; hence, no theoretical differences exist.
On the contrary, the binary relaxations cannot be directly extended for multi-class classification.
Similarly to the binary scenario, two relaxation are available for the multi-class scenario:
\begin{itemize}
    \item Diagonal Equality, which focuses on equal Recall (\textit{i.e.,} $\hat{Y}=Y$) across groups;
    \item Off-diagonal Equality, which focuses on equal off-diagonal misclassification rates (\textit{i.e.} $\hat{Y} \neq Y$) across groups.
\end{itemize}

\subsubsection{Relaxed Separation Criteria}
\label{sec:sep_relaxations}
EO mandates strict conditions to be satisfied; thus, scholars developed relaxations that are easier to satisfy.
In the binary setting, \gls{eo} relaxations restrict equal probability constraints to individual confusion matrix columns $Y$ overt the confusion matrices constructed for each groups.
Extending the probabilistic metric definitions to the multi-class domain breaks this column-wise alignment.
Consequently, Diagonal Equality isolates the main diagonal entries ($n_{kk}$), conversely, Off-diagonal Equality constrains all off-diagonal entries ($n_{kj}$ where $k \neq j$).
Together, these two multi-class relaxations partition the full confusion matrix, such that satisfying both simultaneously is mathematically equivalent to achieving full multi-class Separation.

\label{sec:equal_opportunity}
Equal Opportunity is the first binary relaxation proposed in the literature, which enforces equal \gls{tpr} across protected groups~\cite{mehrabininareh10galstyan,pessach2022review,rabonato2025systematic,castelnovo2022clarification}:
\begin{equation}
    P(\hat{Y} = 1 \mid A = a, Y = 1) = P(\hat{Y} = 1 \mid A =b, Y = 1) \quad \forall a,b \in \mathcal{A}.
\end{equation}
In the binary setting, enforcing Equal Opportunity across protected groups inherently yields equal complementary rates (\textit{i.e.}, \gls{fnr}).

Predictive Equality complements Equal Opportunity by enforcing equal \gls{fpr} across protected groups~\cite{castelnovo2022clarification}:
\begin{equation}
    P(\hat{Y} = 1 \mid A = a, Y = 0) = P(\hat{Y} = 1 \mid A =b, Y = 0) \quad \forall a,b \in \mathcal{A}.
\end{equation}
In this formulation, enforcing Predictive Equality across protected groups inherently requires equal complementary rates (\textit{i.e.}, \gls{tnr}).

Existing literature uses inconsistent terms when referring to multi-class \gls{eo} relaxation, such as \textit{multi-class Equal Opportunity}, \textit{Overall Accuracy},  \textit{Overall Misclassification Rate} or \textit{Max Loss Discrepancy}.
To resolve this ambiguity, we propose the taxonomy \textit{Diagonal Equality} and \textit{Off-diagonal Equality}, whose definitions are given below.

Diagonal Equality is the first multi-class relaxation, and it requires the predictor to score equal recall across every protected group.
In other words, it requires the confusion matrices of all protected groups to have identical diagonals.
Earlier work by Rouzot \textit{et al.} presented a multi-class generalization named Equal Opportunity, also known as \textit{Overall Accuracy}~\cite{rouzot2022learning}.
Although similar to our notion, the authors limited their definition of the metric to a single binary sensitive attribute.
Conversely, Purificato \textit{et al.}~\cite{purificato2024toward}, proposed the following complete multi-class and multi-group formulation:
\begin{equation}
\label{eq:diag_eq}
    P(\hat{Y}=k | A=a, Y=k) = P(\hat{Y}=k | A=b, Y=k), \quad \forall k \in \mathcal{Y}, \forall a, b \in \mathcal{A}.
\end{equation}
Note that in their work, Purificato \textit{et al.} state that Equation~\ref{eq:diag_eq} would also apply as the extension of  EO in the multi-class and multi-group setting.
In contrast, given Hardt \textit{et al.} definition of Equal Opportunity~\cite{hardt2016equality} (see Equation~\ref{eq:eq_odds}), it is clear that the two probabilistic formulations differ, where Equation~\ref{eq:diag_eq} does not fully enforce Separation Criterion across the whole confusion matrix.

Finally, Off-diagonal Equality mandates equal misclassification rates across protected groups.
Rouzot \textit{et al.} described a multi-class generalization named \textit{Overall Misclassification Rate}, also known as \textit{Max Loss Discrepancy}~\cite{rouzot2022learning}, defined as the complementary probability of Diagonal Equality:
\begin{equation}
    P(\hat{Y} \neq j \mid A=a, Y=j) = P(\hat{Y}\neq j \mid A=b, Y=j), \quad \forall j \in \mathcal{Y};\forall a, b \in \mathcal{A}.
    \label{eq:rouzot_missclassification_rate}
\end{equation}
Analogously, they limited their definition of the metric to a single binary sensitive attribute.
This definition suffers from aggregation, which is a critical limitation similar to the EO \textit{vs} Equal Accuracy scenario, where Equation~\ref{eq:rouzot_missclassification_rate} only checks for overall misclassification rates, loosing information regarding potential systematic disparities.
We tackle this gap in the literature by providing the following multi-group and multi-class generalization named Off-diagonal Equality:
\begin{equation}
    P(\hat{Y}=k \mid A=a, Y=j) = P(\hat{Y}=k \mid A=b, Y=j), \quad \forall k,j \in \mathcal{Y}; j \neq k,\forall a, b \in \mathcal{A}.
    \label{eq:off-diagonal-parity}
\end{equation}
\begin{proposition}[Equivalence to Multi-Class Equalized Odds]
\label{prop:EquivalencetoMulti-ClassEqualizedOdds}
    Satisfying Eq.~\ref{eq:off-diagonal-parity} is mathematically equivalent to satisfying full multi-class \gls{eo} (parity across the entire conditional confusion matrix).
    (See Proof~\ref{proof:ProofEquivalencetoMulti-ClassEqualized Odds} in Appendix~\ref{app:secB}.)
\end{proposition}
We stress that this implication only holds when Off-diagonal Equality is strictly satisfied across all targets and protected groups, a condition rarely met in real world scenarios.
Hence, introducing this relaxation remains highly relevant as it evaluates purely off-diagonal disparities.

\subsection{Rebuilding from the Criteria: Sufficiency}
\label{subsec:rebuild_Sufficiency}
Sufficiency is the third group fairness criterion defined by Barocas \textit{et al.}~\cite{barocas2023fairness}.
It dictates that the ground truth $Y$ must be independent of any sensitive attribute $A$, conditioned on the prediction $\hat{Y}$, \textit{i.e.}
$
    Y \independent A \mid \hat{Y}.
$
Like Separation, Sufficiency is a ``conservative'' fairness criterion, meaning that it is always satisfied by a ``perfect classifier'' ($\hat{Y} = Y$).
As a consequence, if the ground truth is  affected by historical or societal bias, enforcing Sufficiency may perpetuate, rather than mitigate, such biases.

Unlike Independence and Separation, which adopt the point of view of the individual being evaluated, Sufficiency reflects the perspective of the decision maker.
It requires that predictions have the same reliability across protected groups.
Although Separation requires similar confusion matrices across protected groups, 
resulting in equal error rates, Sufficiency requires that individuals receiving the same prediction have the same probability of belonging to the corresponding true class, regardless of their sensitive attribute.

Example~\ref{example:reliability_vs_sensitivity} in Appendix~\ref{app:secA} highlights the difference between Separation and Sufficiency.
Ultimately, Sufficiency is particularly desirable in applications where the reliability of predictions should not depend on sensitive attributes, such as clinical decision support and risk assessment systems.

\subsubsection{Predictive Parity}
Although Predictive Parity is restricted to $Y =\hat{Y}$ and does not fully satisfy Sufficiency for off-diagonal predictions ($Y \neq \hat{Y}$), it provides coverage for the most most desirable practical guarantee.
When decision makers require reliability, the most desirable guarantee is the correctness of positive predictions (\textit{e.g.}, the patient is sick, the loan can be approved); hence, the Sufficiency criterion can be restricted to \glspl{ppv}.
Originally, scholars provided only the binary interpretation for Predictive Parity~\cite{castelnovo2022clarification}:
\begin{equation}
    P(Y=1 \mid  A=a, \hat{Y}=1) = P(Y=1 \mid  A=b, \hat{Y}=1), \quad \forall a, b \in \mathcal{A}.
\end{equation}
Notice that this definition only covers $Y = \hat{Y} = 1$ and ignores negative predictions ($Y = \hat{Y} = 0$), representing a relaxation of the full Sufficiency criterion, which would also require equal Negative Prediction Values.

To the best of our knowledge, scientific literature only covers Sufficiency in binary scenarios.
In this work, we address this gap by proposing a multi-class and multi-group metric extension for Predictive Parity as formalized as follow:
\begin{equation}
    P(Y=k \mid  A=a, \hat{Y}=k) = P(Y=k \mid  A=b, \hat{Y}=k), \quad \forall k \in \mathcal{Y}, \forall a, b \in \mathcal{A}.
\end{equation}

Notice that our multi-class formulation does not generalise the binary definition.
As previously stated, the original  Predictive Parity restrict equality strictly to $Y = \hat{Y} = 1$ assuming a binary scenario. 
In contrast, our multi-class formulation extends this equal-precision requirement across every label $k \in \mathcal{Y}$.
When evaluated on a binary problem, our method enforces two simultaneous conditions:
\begin{equation}
    \begin{aligned}
        P(Y=0 \mid  A=a, \hat{Y}=0) &= P(Y=0 \mid  A=b, \hat{Y}=0) \\
        P(Y=1 \mid  A=a, \hat{Y}=1) &= P(Y=1 \mid  A=b, \hat{Y}=1), \quad \forall a, b \in \mathcal{A}
    \end{aligned}
\end{equation}
which results in a strictly stronger requirement than the standard binary definition.

\subsection{Rebuilding from the Criteria: Accuracy}
\label{subsec:rebuild_Accuracy}
Although Barocas \textit{et al.}~\cite{barocas2023fairness} proposed only three criteria (Independence, Separation, Sufficiency), algorithmic fairness literature covers a fourth condition: Accuracy.
Formally, Accuracy does not constitute a distinct criterion; mathematically, requiring equal overall accuracy across protected groups does not align with or imply the metrics associated with the three well-established criteria.

We stress once again the difference between requiring equal overall accuracy across protected groups and enforcing Separation (or Sufficiency), where 
evaluating accuracy alone as a fairness metric  can be misleading and may hide severe \gls{ml} prejudices (see Example \ref{example:eq_odds_vs_eq_acc_repayingLoan} in Appendix~\ref{app:secA}).
Specifically, Accuracy criteria evaluates unconditional total correctness and restricts only the aggregate proportion of decisions, permitting the model to commit vastly different types of errors across protected groups.
Moreover, alongside Separation and Sufficiency, Accuracy is considered a ``conservative'' criteria because it is trivially satisfied by a ``perfect'' classifier ($\hat{Y}=Y$).

\subsubsection{Equal Accuracy}
Equal Accuracy, also known as \textit{Accuracy Parity} or \textit{Overall Accuracy Equality}, is a group fairness metric that, as the name suggests, requires overall prediction accuracy across protected groups.
Castelnovo \textit{et al.}~\cite{castelnovo2022clarification} covered the binary setting, which we consequently extend as follows:
\begin{equation}
    P(Y=\hat{Y} \mid A=a) = P(Y=\hat{Y} \mid A=b), \quad \forall a, b \in \mathcal{A}.
    \label{eq:eq_acc}
\end{equation}
Mathematically, the binary and multi-class formulations of Equal Accuracy are structurally equivalent: both compute the total probability mass along the diagonal of the group-specific confusion matrix normalized by total group size.
Once more, the binary scenario is simply a special two-class case ($\lvert\mathcal{Y}\rvert=2$).
Purificato \textit{et al.} propose multi-class and multi-group Overall Accuracy Equality~\cite{purificato2024toward}:
\begin{equation}
    \sum_{k\in \mathcal{Y}}{ P(Y=k \mid  A=a, \hat{Y}=k)} =  \sum_{k\in \mathcal{Y}}{P(Y=k \mid  A=b, \hat{Y}=k)}, \quad \forall a, b \in \mathcal{A}.
\end{equation}
Note that, despite the similar name, this formulation differs from Equation~\ref{eq:eq_acc} because it sums the group-wise precisions.
Formally, because Overall Accuracy Equality conditions on the predicted label $\hat{Y}$, this metric aligns with the Sufficiency criteria (evaluating predictive reliability) rather than Accuracy criteria.
Ultimately, similarly to Equal Accuracy, Overall Accuracy Equality suffers relevant loss of information regarding systematic behaviour, due to aggregation produced by the sum; we thus encourage the use of multi-class Predictive Parity.

\section{Mutual Exclusivity and Coexistence of Observational Fairness Criteria}
\label{sec:mutal_exclusion}
The foundational analysis by Barocas \textit{et al.}~\cite{barocas2023fairness} established that key observational fairness criteria -- namely Independence, Separation, and Sufficiency -- are inherently incompatible in many real-world classification settings.
The authors propose cardinality-agnostic proofs (\textit{i.e.,} $\lvert\mathcal{Y}\rvert > 2$) of  \textit{Independence vs. Sufficiency} and \textit{Separation vs. Sufficiency} under strictly positive joint distributions.
Furthermore, they prove the mutual exclusivity  of \textit{Separation vs. Sufficiency} under the binary classification task, and further weakened their assumption to prove that an imperfect classifier cannot simultaneously satisfy \textit{Separation} and \textit{Sufficiency} in the limited binary scenario.
In the remainder of this Section, we extend the propositions to the multi-class scenario, and more notably, we state that Independence and Separation \emph{can} coexist under strict conditions (the proofs are shown in Appendix~\ref{app:secB}).

\subsection{Mutual Exclusivity of Independence and Separation}
To avoid notational ambiguity, we explicitly distinguish between scalar probabilities and their vector representations. 
Let $P(Y=y)$ denote the scalar marginal probability of the true label being $y$. 
We define the column vector $\mathbf{p}_Y \in \mathbb{R}^k$ such that its $y$-th component is $P(Y=y)$.
Similarly, let $P(Y=y \mid A=a)$ be the conditional probability, represented by the column vector $\mathbf{p}_{Y \mid a} \in \mathbb{R}^k$.
Let $\mathbf{p}_{\hat{Y}} \in \mathbb{R}^{k}$ and $\mathbf{p}_{\hat{Y} \mid a} \in \mathbb{R}^{k}$ be the corresponding marginal and conditional probability vectors for the predictions.

We define the demographic disparity vector for a given group $a \in \mathcal{A}$ as
$
    \Delta \mathbf{p}_a = \mathbf{p}_{Y \mid a} - \mathbf{p}_Y
$
which implies that for each element $y$, the disparity is $\Delta p_{y \mid a} = P(Y=y \mid A=a) - P(Y=y)$.
Let $M$ be a column-stochastic matrix of size $k \times k$ representing the classifier's conditional probabilities, where each element is defined as $M_{j,y} = P(\hat{Y}=j \mid Y=y)$.
We establish the following base premises for our classification problem: \textit{(i)} $A \not\perp Y$, \textit{i.e.}, there exists at least one group $a \in \mathcal{A}$ such that $\Delta \mathbf{p}_a \neq \mathbf{0}$, and \textit{(ii)} $\hat{Y} \not\perp Y$. 

\begin{proposition}[Mutual Exclusion of Independence and Separation]
\label{prop:indep_sep_mutual_exclusion}
    Under the base premises for a multi-class setting ($k > 2$), if the demographic disparity vector does not lie within the null space of the classifier matrix \big($\Delta \mathbf{p}_a \notin \text{Kernel}(M)$\big), then the fairness criteria of Independence ($\hat{Y} \perp A$) and Separation ($\hat{Y} \perp A \mid Y$) are mutually exclusive. (See Proof~\ref{proof:MutualExclusionofIndependenceandSeparation} in Appendix~\ref{app:secB}.)
\end{proposition}
\subsection{Coexistence of Independence and Separation}
\label{sec:indep_sep_coexistnce}
In contrast to the binary classification tasks, Independence and Separation can mathematically coexists under strict conditions in multi-class scenarios.

\begin{proposition}[Coexistence of Independence and Separation]
\label{prop:CoexistenceofIndependenceandSeparation}
Under the base premises for a multi-class setting ($k > 2$), Independence ($\hat{Y} \perp A$) and Separation ($\hat{Y} \perp A \mid Y$) can hold simultaneously if and only if the classifier's conditional probability matrix $M$ has a rank $r \in [2, k-1]$ and every demographic disparity vector $\Delta \mathbf{p}_a$ lies entirely within the null space of $M$ (\textit{i.e.}, $\Delta \mathbf{p}_a \in \text{Kernel}(M)$ for all $a \in \mathcal{A}$).
(See Proof~\ref{proof:CoexistenceofIndependenceandSeparation} in Appendix~\ref{app:secB}.)
\end{proposition}

While mathematically sound, satisfying these requirements in real-world applications is exceedingly rare due to a fundamental decoupling between data-driven disparities and model dynamics.
On the one hand, the demographic disparity vector $\Delta \mathbf{p}_a$ reflects intrinsic joint distributions of the underlying population ($A \not\perp Y$) and remains immutable unless explicitly modified through data pre-processing techniques.
On the other hand, the confusion matrix $M$ is highly sensitive and non-deterministic; it depends not only on the specific task and dataset, but also on model architecture, optimization choices, and run-specific stochasticity (\textit{e.g.}, random weight initialization and mini-batch shuffling).
Consequently, expecting a non manipulated data disparity vector to naturally align with the null space of a highly volatile, run-dependent classifier matrix is seldom achievable without targeted algorithmic interventions.

\subsection{General Mutual Exclusivity of Separation and Sufficiency}
Hardt \textit{et al.}~\cite{barocas2023fairness} provided a formal proof the mutual exclusivity between Separation and Sufficiency under the assumption of strictly positive probabilities for all events in the joint distribution of $(A, \hat{Y},Y)$.
In later formulations, they weakened this assumption to prove mutual exclusivity, yet remained limited to binary classification.

We bridge this literature gap providing a general proof of their mutual exclusivity, weakening the original conditions to the existence of a single class $j \in \hat{\mathcal{Y}}$ and sensitive group $a \in \mathcal{A}$ with strictly positive  probabilities for the conditional events $P(\hat{Y}=j \mid A=a) > 0$ and $P(\hat{Y}=j \mid Y=k) > 0 \forall k \in \mathcal{Y}$.

We assume the following base conditions: \textit{(i)} $A \not\perp Y$, hence $\exists a \in \mathcal{A}, k \in \mathcal{Y}$ such that $P(Y=k \mid A=a) \neq P(Y=k)$, and \textit{(ii)} given $a \in \mathcal{A}$ which invalidates the statistical independence of $A$ and $Y$, there exists a $j \in \hat{\mathcal{Y}}$ such that $P(\hat{Y}=j \mid A=a) > 0$ and $P(\hat{Y}=j \mid Y=k) > 0 \forall k \in \mathcal{Y}$.
Note that that this implies that $\hat{Y}$ cannot be a perfect classifier.

\begin{proposition}[Mutual Exclusion of Separation and Sufficiency]
\label{thm:sep_vs_suff}
Under these base conditions, the fairness criteria of Separation ($\hat{Y} \perp A \mid Y$) and Sufficiency ($Y \perp A \mid \hat{Y}$) cannot hold simultaneously.
(See Proof~\ref{proof:MutualExclusionofSeparationandSufficiency} in Appendix~\ref{app:secB}.)
\end{proposition}


\section{Experimental Evaluation and Results}
\label{sec:experimental}

All results in this manuscript are based on  \acro \footnote{\url{https://github.com/FRACTALS-Research-Group/FairMetrics}}~\cite{fairmetricsrepo}, an open-source PyTorch-based Python library for binary, multi-class, and multi-group fairness evaluation. 
We implemented in \acro all metrics discussed in this manuscript.
The codebase comprises two main packages:
\textit{(i)}~\texttt{fairmetrics/}, the core evaluation engine containing discrete fairness metrics functions organized by foundational criteria; and \textit{(ii)}~\texttt{fairmetrics\_experiments/}, the experimental suite containing data loaders, target binarization strategies, and the \gls{hpo} evaluation runner.
The runner supports native multi-class evaluation alongside binarization reductions to quantify the empirical ``reduction gap'', with all experiments fully reproducible via a single \texttt{batch} script exporting structured \texttt{JSON} results.

In order to compare binary and multi-class/multi-group metrics, we applied \acro to two open datasets for fairness in classification, as described in the next subsection.

\subsection{Experimental Setup}
We evaluated our metrics on two widely established tabular datasets in the fair \gls{ml}: \textit{Adult}~\cite{adult_2} and \textit{Students' Dropout and Academic Success}~\cite{martins2021early}.
Both datasets are designed for the classification tasks, and, in their original formulations, are characterized by class imbalance and sensitive demographic features.
Currently, the literature provides few real-world multi-class datasets for fairness evaluation and fewer studies regarding multi-class fairness approaches to align with.
To maintain realistic feature distributions without resorting to synthetic generation, we created a custom multi-class target for the \textit{Adult} dataset by categorizing individuals into three classes based on income and education level.
See Exploratory Data Analysis in Appendix~\ref{app:secC}.
All experiments were executed on a virtualized \texttt{Ubuntu 24.04 LTS} 23-node cluster (Intel Xeon Gold / AMD EPYC CPUs, up to 400~GB RAM) and a dedicated \gls{vm} (12 vCPUs AMD EPYC, 94~GiB RAM) backed by \gls{nas}.
Throughout our experiments, in order to ensure reproducibility we set random seeds to $42$ for the initial $10$-fold stratified split (\texttt{StratifiedKFold}) and the fold index $k \in \{0, \dots, 9\}$ for fold-level \gls{rf} models and \gls{smotenc} oversampling.

\subsection{Evaluation Methodology}
In current algorithmic fairness literature, standard methodologies and metrics primarily target binary classification, frequently avoiding any multi-class discussion.
To overcome these limitation, practitioners might resort to two main One-vs-Rest reduction approaches to decompose a fair multi-class classification task into a set of binary classification sub-problems, discussed below.

The first approach is called Target Binarization and partitions the label set $\mathcal{Y}$ into two subsets, mapping the original multi-class target $K = \lvert \mathcal{Y}\rvert$ into a single synthetic binary target.
Given the new binary target, any preferred optimization strategy can be used to train 
a binary classifier, so that the model's fairness can be evaluated using any existing binary metric.
This approach has multiple limitations:
\textit{(i)} forced binarization of multi-class labels discards a lot of information;
\textit{(ii)} class aggregation dramatically alters the classification task and distorts the the semantic meaning of each label, which
\textit{(iii)} potentially exacerbates representation imbalances within disadvantaged protected groups, hence
\textit{(iv)} masks hidden disparities which could harm specific under-represented subgroups;
ultimately \textit{(v)} selecting an appropriate binary partition is difficult when multiple classes characterize high-stake scenarios.

The second approach is called Prediction Binarization trains a multi-class classifier on the original multi-class dataset, and then applies prediction binarization only during fairness evaluation.
While this method keeps the original multi-class nature of the problem, and results in a more useful multi-class classifier, it does not fully resolves the previous limitations inherent to binary reduction.

Not all fairness metrics defined in this manuscript are directly comparable to each other due to the differences in their mathematical formulations.
Notably, the binary and multi-class relaxation of the Separation criteria require the equality of different partitions of the confusion matrix, 
and thus a comparison is not relevant in this scenario.
Similarly, Binary Predictive Parity, which refers to the Sufficiency criteria, requires equal \glspl{ppv} for the positive prediction $\hat{Y}=1$, whereas its multi-class extension enforces equal \glspl{ppv} for every possible label.
Despite this difference, we decide to compare Predictive Parity to highlights the limitation of the binary metric when applied to multi-class outcomes.

We evaluate both reduction paradigms across all candidate One-vs-Rest partitions to provide the full coverage of all possible behaviours.
We choose a Random Forest classifier (\gls{rf}) as our base \gls{ml} model given its robustness, high effectiveness with tabular data and computational efficiency during repeated cross-validation.
As a first step, we train the multi-class \gls{rf} on the  original dataset using $10-$fold \gls{cv}. 
The fairness is evaluated using our proposed multi-class and multi-group fairness metrics.
Subsequently, we average the fairness scores across folds to provide a more robust evaluation.
This final score of this Native Multi-Class Model is herein denoted as $\text{MC}_{\text{Native}}$, unambiguously denoting the ground-truth multi-class training and evaluation baseline.

As second stage, the multi-class predictions are mapped into One-vs-Rest binary targets, where the selected class is assigned to the positive label ($1$) while all remaining classes are mapped to the negative label ($0$); subsequently, we evaluate binary fairness metrics, and we average the fairness scores across across the $10$ folds.
This final score is herein denoted as $\text{MC}_{\text{OvR}}$, denoting that the model is multi-class, but its outputs are binarized \textit{post-hoc} One-vs-Rest for evaluation.

Finally, we apply the same mappings of the previous approach to the original ground truth and train binary \glspl{rf} on the One-vs-Rest binary targets.
Similarly, we average the fairness scores across the $10$ folds.
This final score is herein denoted as $\text{Bin}_{\text{OvR}}$, denoting binary model trained on \textit{Ex-Ante} binarized targets.

To quantify the empirical bias introduced by each reduction paradigm relative to the native multi-class baseline $\text{MC}_{\text{Native}}$, we define the pre-training and post-hoc evaluation gaps as:
\begin{equation}
    \Delta_{\text{Pre}} = \text{Bin}_{\text{OvR}} - \text{MC}_{\text{Native}} \qquad \text{and} \qquad \Delta_{\text{Post}} = \text{MC}_{\text{OvR}} - \text{MC}_{\text{Native}},
\end{equation}
where $\Delta_{\text{Pre}}$ isolates the disparity shift induced by the binarization of targets before model training, and $\Delta_{\text{Post}}$ isolates the distortion introduced by post-hoc prediction binarization during evaluation.
These quantities range within $[-1, 1]$, where a negative gap indicates that the reduction approach underestimates true disparity, and a positive gap indicates overestimation.
Furthermore, in the Results section, negative gaps are highlighted in red hues, while positive gaps are highlighted in blue hues.

For each dataset and binary mappings, we first performed \gls{hpo} via $10-$fold \gls{cv} grid search, maximizing the Macro $F_1$ Score.
The search space for the \gls{rf} classifier included: the number of estimators ($200, 400, 600$), maximum depth ($12, 15, 18, 20$, or \texttt{None}), minimum samples split ($5, 10, 15, 20$), minimum samples leaf ($1, 2, 3$), maximum features ($0.4, 0.5, 0.6$, or \texttt{'sqrt'}), and class weight configurations (\texttt{'balanced'} or \texttt{'balanced\_subsample'}).

\subsection{Empirical Analysis and Discussion}
Given the extensive empirical evaluation conducted, this Section summarizes the core insights, general considerations and emerged behaviours observed across our experiments, whereas the complete result analysis is accessible in the Appendix~\ref{app:secD}.

To enrich our discussion and to provide concrete examples which comprise most of the relevant considerations: Table~\ref{tab:t3_fairness_sex} provides the results of the fairness evaluation using our \textit{Adult Multi-class} dataset conducted selecting Sex as the sensitive feature, whereas Table~\ref{tab:t1_fairness_gender_x_international} presents the results \textit{Students’ Dropout and Academic Success} where Gender $\times$ International is the sensitive feature.
Despite showing a small but significant subset of empirical experiments, we stress that throughout all the tested scenarios, $\Delta_{\text{Pre}}$ and  $\Delta_{\text{Post}}$ consistently score non-negligible gaps.
\begin{table}[!htbp]
\centering

\renewcommand{\arraystretch}{0.85}
\begin{tabularx}{\textwidth}{X c c c G G}
\toprule
\textbf{Fairness Metric} & \textbf{$\text{Bin}_{\text{OvR}}$} & \textbf{$\text{MC}_{\text{OvR}}$} & \textbf{$\text{MC}_{\text{Native}}$} & \multicolumn{1}{c}{\textbf{$\Delta_{\text{Pre}}$}} & \multicolumn{1}{c}{\textbf{$\Delta_{\text{Post}}$}} \\
\midrule
\multicolumn{6}{l}{\textbf{Target 0: Low-Income \& Low-Education}} \\
\midrule
Demographic Parity & $0.0236$ & $0.0487$ & $0.3228$ & $-0.2993$ & $-0.2741$ \\
Conditional Demographic Parity & $0.0470$ & $0.0777$ & $0.3327$ & $-0.2857$ & $-0.2551$ \\
Equalized Odds & $0.0696$ & $0.0914$ & $0.3342$ & $-0.2646$ & $-0.2428$ \\
Predictive Parity & $0.0196$ & $0.0375$ & $0.0650$ & $-0.0454$ & $-0.0274$ \\
Equal Accuracy & $0.0184$ & $0.0163$ & $0.0368$ & $-0.0184$ & $-0.0204$ \\
\midrule
\multicolumn{6}{l}{\textbf{Target 1: Low-Income \& High-Education}} \\
\midrule
Demographic Parity & $0.3449$ & $0.3228$ & $0.3228$ & $+0.0221$ & $0.0000$ \\
Conditional Demographic Parity & $0.3538$ & $0.3327$ & $0.3327$ & $+0.0211$ & $0.0000$ \\
Equalized Odds & $0.3361$ & $0.3342$ & $0.3342$ & $+0.0019$ & $0.0000$ \\
Predictive Parity & $0.0646$ & $0.0462$ & $0.0650$ & $-0.0003$ & $-0.0188$ \\
Equal Accuracy & $0.0349$ & $0.0375$ & $0.0368$ & $-0.0019$ & $+0.0007$ \\
\midrule
\multicolumn{6}{l}{\textbf{Target 2: High-Income}} \\
\midrule
Demographic Parity & $0.2611$ & $0.2741$ & $0.3228$ & $-0.0617$ & $-0.0487$ \\
Conditional Demographic Parity & $0.2714$ & $0.2853$ & $0.3327$ & $-0.0613$ & $-0.0474$ \\
Equalized Odds & $0.1564$ & $0.1778$ & $0.3342$ & $-0.1778$ & $-0.1564$ \\
Predictive Parity & $0.0326$ & $0.0388$ & $0.0650$ & $-0.0324$ & $-0.0262$ \\
Equal Accuracy & $0.1240$ & $0.1248$ & $0.0368$ & $+0.0872$ & $+0.0881$ \\
\bottomrule
\end{tabularx}
\caption{Fairness Metrics Comparison for Sensitive Feature: Sex (\textit{Adult Multi-class} Dataset)}
\label{tab:t3_fairness_sex}
\end{table}

\begin{table}[!htbp]
\centering

\renewcommand{\arraystretch}{0.85}
\begin{tabularx}{\textwidth}{X c c c G G}
\toprule
\textbf{Fairness Metric} & \textbf{$\text{Bin}_{\text{OvR}}$} & \textbf{$\text{MC}_{\text{OvR}}$} & \textbf{$\text{MC}_{\text{Native}}$} & \multicolumn{1}{c}{\textbf{$\Delta_{\text{Pre}}$}} & \multicolumn{1}{c}{\textbf{$\Delta_{\text{Post}}$}} \\
\midrule
\multicolumn{6}{l}{\textbf{Target 0: Dropout}} \\
\midrule
Demographic Parity & $0.3208$ & $0.3922$ & $0.4837$ & $-0.1629$ & $-0.0915$ \\
Conditional Demographic Parity & $0.5553$ & $0.5913$ & $0.7090$ & $-0.1538$ & $-0.1177$ \\
Equalized Odds & $0.3263$ & $0.4313$ & $0.6139$ & $-0.2877$ & $-0.1826$ \\
Predictive Parity & $0.2349$ & $0.3753$ & $0.6331$ & $-0.3982$ & $-0.2578$ \\
Equal Accuracy & $0.1651$ & $0.1805$ & $0.2428$ & $-0.0777$ & $-0.0623$ \\
\midrule
\multicolumn{6}{l}{\textbf{Target 1: Enrolled}} \\
\midrule
Demographic Parity & $0.2380$ & $0.2664$ & $0.4837$ & $-0.2457$ & $-0.2173$ \\
Conditional Demographic Parity & $0.3161$ & $0.4197$ & $0.7090$ & $-0.3929$ & $-0.2894$ \\
Equalized Odds & $0.5573$ & $0.4456$ & $0.6139$ & $-0.0566$ & $-0.1683$ \\
Predictive Parity & $0.3497$ & $0.5281$ & $0.6331$ & $-0.2834$ & $-0.1051$ \\
Equal Accuracy & $0.4037$ & $0.2808$ & $0.2428$ & $+0.1609$ & $+0.0380$ \\
\midrule
\multicolumn{6}{l}{\textbf{Target 2: Graduate}} \\
\midrule
Demographic Parity & $0.4537$ & $0.4140$ & $0.4837$ & $-0.0300$ & $-0.0697$ \\
Conditional Demographic Parity & $0.6691$ & $0.5833$ & $0.7090$ & $-0.0399$ & $-0.1257$ \\
Equalized Odds & $0.4794$ & $0.3979$ & $0.6139$ & $-0.1345$ & $-0.2161$ \\
Predictive Parity & $0.4667$ & $0.3276$ & $0.6331$ & $-0.1664$ & $-0.3055$ \\
Equal Accuracy & $0.3421$ & $0.1743$ & $0.2428$ & $+0.0993$ & $-0.0685$ \\
\bottomrule
\end{tabularx}
\caption{Fairness Metrics Comparison for Sensitive Feature: Gender $\times$ International (\textit{Students’ Dropout and Academic Success} Dataset, Original Setup)}
\label{tab:t1_fairness_gender_x_international}
\end{table}

First and most critically, our results demonstrate that both the One-vs-Rest reduction techniques consistently exhibit relevant fairness expectations gaps across all evaluated metrics.
Furthermore, within a single target mapping, multiple fairness metrics can simultaneously yield alarming discrimination gaps, while others report deceptively negligible disparities.
Throughout our evaluation, we categorize these discrepancies into three possible behaviours: underestimation, overestimation, and negligible gap.

Most frequently, the One-vs-Rest techniques results being optimistic filters, which systematically under-report the true fairness disparities across protected groups by up to $-0.4$.
This behaviour represents the most relevant criticality of binary reduction.
Furthermore, aggregating together distinct classes proved to potentially mask prominent treatment disparities, which is extremely exacerbated when combining under-represented classes or protected groups.
Although less common, unfairness is overestimated in some binary mapping scenarios, peaking at a significant $+0.1609$ for Equal Accuracy on the \textit{Target 1: Enrolled} mapping ($\Delta_{\text{Pre}}$ in Table~\ref{tab:t1_fairness_gender_x_international}).
This overestimation occurs because binarizing a minority class (\textit{Enrolled}, $\sim 18\%$) against a merged majority category (\textit{Dropout} and \textit{Graduate}, $\sim 82\%$) forces the binary classifier to prioritize the negative majority class (achieving $91.65\%$ Specificity versus $41.80\%$ Recall, as shown in Table~\ref{tab:t1_rf_performance} in Appendix~\ref{app:secD}).
Consequently, group-wise variations in true negative rates and subgroup base-rate compositions on the composite majority category artificially inflate the overall binary accuracy disparity between demographic groups ($\text{Bin}_{\text{OvR}} = 0.4037$), whereas native multi-class evaluation assesses true three-class diagonal performance ($\text{MC}_{\text{Native}} = 0.2428$).
This behaviour could exacerbate existing biases in datasets or predictions, potentially leading to harmful consequences in real-world, high-stakes scenarios.
Moreover, it could mislead practitioners into concluding that the \gls{ml} model or dataset is biased, when, in reality, the original multi-class distribution and predictions are fair.

Even when a specific binary mapping achieves near-zero gaps, it is not possible to determine \textit{a-priori} whether this reflects true equity or a reduction artifact.
$\Delta_{\text{Pre}}$ and $\Delta_{\text{Post}}$ gaps approach zero only in the rare event that a binary mapping happenstance isolates the specific class exhibiting the greatest disparities among protected groups, as showcased in the Low-Income \& High-Education vs rest scenario, when comparing sex disparities in the \textit{Adult Multi-class} Dataset.
This is solely possible when a single class is affected by systematic discrimination, a condition that is not satisfied in many real-world scenarios and is most likely is unknown \textit{a priori}.
Finally, we stress that the metrics discussed in this work evaluate disparities among empirical estimations of probabilities, providing no information about or insight into the true underlying behaviour.
Hence, a near-null fairness estimation or gap provides no insurance of the absence of any complex or hidden discriminatory behaviour.
Importantly, different One-vs-Rest reduction techniques might produce noticeable different behaviour, on the same dataset.
Table~\ref{tab:t1_fairness_gender_x_international} showcase a striking example where Equal Accuracy is simultaneously overestimated ($\Delta_{\text{Pre}}$ reaches $+0.0993$) and underestimated ($\Delta_{\text{Post}}$ scores $-0.0685$).
Furthermore, our experiments demonstrate that these binary reduction techniques are not suitable for class-imbalanced datasets.
Merging classes frequently masks minority group dynamics within the majority population, skewing probabilities disparities.
Lastly, fairness-unaware oversampling procedures such as \gls{smotenc}, while generally effective for improving overall predictive performance, fail to reduce the issues described above.
In this regard, several pre-processing fairness-aware techniques have been proposed in literature and are particularly well suited to this task.

Our empirical evaluation highlights the issues of relying on One-vs-Rest reduction techniques for fairness assessment.
We cover all the comparable binary and multi-class multi-group fairness metrics and demonstrate that no \textit{a priori} assumption can be made regarding their agreement; therefore, this binary reduction techniques offer an unreliable foundation for algorithmic \gls{ai} fairness auditing.
Consequently, in a multi-class classification scenario, fairness evaluation must be directly conducted using native multi-class and multi-group fairness metrics, such as those proposed in this work.

From a legal and ethical point of view, it has to be noted that experimental results bear on a question that the technical literature usually treats as a matter of convenience: how the decision on which fairness is assessed is represented. Reducing a K-class decision to one or more binary problems is not a neutral computational step but a classificatory act that redefines the outcome categories and, with them, the term of comparison against which group disadvantage is measured. The classification of persons and outcomes is itself an exercise of normative power, whose legitimacy must be justified. However, the law has yet to embed these considerations into its principles: under the AI Act, for example, the examination of possible biases required by Article 10(2)(f)–(g), and its documentation under Article 11, may be formally discharged on the basis of a binary assessment, as it specifies neither the label space nor the metrics on which bias is to be examined. Under EU non-discrimination law, statistical evidence of disparity may establish facts from which indirect discrimination is presumed, shifting the burden of proof to the respondent (Article 8 of Directive 2000/43/EC).

\section{Conclusion}
\label{sec:conclusion}
In this work, we address a fundamental gap in algorithmic fairness: the lack of
native multi-class and multi-group fairness metrics, and the widespread reliance on
binary reduction paradigms.
Standard fairness audits frequently decompose multi-category outcomes into binary
sub-problems (\textit{e.g.}, via \textit{One-vs-Rest} target or prediction binarization); we established that these strategies introduce sever distortion into bias assessments.
To overcome these limitations, we present a theoretical framework structured around Barocas \textit{et al.}'s foundational criteria~\cite{barocas2023fairness} -- namely Independence, Separation, and Sufficiency.
First, we organize the existing literature on multi-class and multi-group fairness and derive the missing formulations needed to construct a complete taxonomy, covering Demographic Parity, Conditional Demographic Parity, Equalized Odds, Predictive Parity, and Equal Accuracy across arbitrary cardinalities, as presented in Table~\ref{tab:fairness_metrics}.
Second, we partition the multi-class conditional confusion matrix to formalize two multi-class Separation relaxations: \textit{Diagonal Equality}, which constrains class-wise Recall across protected cohorts, and \textit{Off-diagonal Equality}, which constrains misclassification rates to isolate off-diagonal disparities.
Third, we adopt a multi-dimensional disparity evaluation operator grounded in the Rawlsian principle of justice, which evaluates worst-case disparities across protected group pairs~\cite{edor2020john}.

Beyond metric design, we deepen the theoretical understanding of observational fairness criteria through formal mathematical proofs that hold for arbitrary label cardinalities. 
Specifically, we prove that satisfying Off-diagonal Equality is mathematically equivalent to  satisfying full multi-class Equalized Odds (Proposition~\ref{prop:EquivalencetoMulti-ClassEqualizedOdds}).
We also prove the exact algebraic conditions governing Independence and Separation, characterizing when these criteria are mutually exclusive and demonstrating that they coexist if and only if demographic disparity vectors belong to the null space of a reduced-rank classifier matrix (Propositions~\ref{prop:indep_sep_mutual_exclusion}, \ref{prop:CoexistenceofIndependenceandSeparation}).
Finally, we prove the general mutual exclusivity of Separation and Sufficiency under non-degeneracy conditions weaker than full joint positivity (Proposition~\ref{thm:sep_vs_suff}).

Empirically, to demonstrate the necessity of adopting native multi-class fairness metrics in multi-class classification scenarios, we conducted extensive empirical experiments comparing the multi-class fairness assessments against evaluations performed using binarization strategies.
Our evaluations on the multi-class \textit{Adult}~\cite{adult_2}, and \textit{Students' Dropout and Academic Success}~\cite{martins2021early} datasets confirm that binary reductions provided unreliable basis for fairness auditing.
Binary reductions predominantly under-report disparities by up to $-0.40$, but can also inflate them when binarizing minority classes due to severe Specificity/Recall trade-offs.

From a legal and ethical perspective, we suggest that the normative choices underlying the development of CS, such as choosing a specific fairness metric, should itself be documented.
Choosing among Independence, Separation and Sufficiency, and within Separation between Diagonal Equality and full Equalized Odds, means choosing among different conceptions of equality.
Article 10(2)(d) of the \gls{euaiact} already requires data-governance practices to include the formulation of assumptions, ``in particular with respect to the information that the data are supposed to measure and represent''.

Lastly, we release \acro, an open-source Python library for assessing fairness in any classification scenario, implementing all proposed metrics.

Despite these contributions, our work presents some limitations.
It focuses only on classification tasks and relies solely on maximum difference aggregation strategy.
These limitations outline promising avenues for future research: we aim to extend our fairness framework by incorporating alternative aggregation operators and extending metric formulations to continuous regression tasks.

\begin{acks}
This work is supported by the high-performance computing infrastructure developed under the project ``CONVECS”, funded by the PR Veneto FESR 2021-2027 program, Priority 1 – Specific Objective 1.1 – Action 1.1.2.
\end{acks}

\section*{Generative AI Usage Statement}
Generative AI tools (Claude Opus 5.5 Pro, Gemini Flash, Gemini Pro, and Gemini Reasoning), accessed through an institutional service, supported manuscript preparation.
Their use was limited to generating LaTeX code for formatting, tables, and figures, as well as providing minor assistance with grammar, syntax, and spelling.
All AI-generated or AI-assisted material was carefully reviewed and verified by the authors, who take full responsibility for the manuscript's accuracy, originality, and scientific integrity.

\section*{Ethics and Privacy Statement}
The authors declare they have fulfilled all required privacy and ethics obligations.

\bibliographystyle{ACM-Reference-Format}
\bibliography{sample-base}

\appendix
\section{Examples}\label{app:secA}
In this Section we provide various illustrative scenarios which highlights the differences and desiderate applications of each fairness criteria and fairness metric.

\begin{example}[Fairness criteria difference: loan approval]
\label{example:loan_approval1}
    A bank uses a \gls{ml} model to approve or deny credit loans to their clients and that gender is selected as the sensitive attribute, considering, for the sake of simplicity, two groups: males and females.
    
    \textit{Independence} ensures the approval rates to be equal across each demographic group, \textit{e.g.} males and females.
    Thus the applicant's gender does not influence the approval rate.

    \textit{Separation} requires that correct classification rates and error rates must be equal for both males and females, thus removing any performance disparities between protected groups.
    Hence, any correlation between the prediction and the client's gender is explained solely by applicant's true creditworthiness.
    
    Lastly, \textit{Sufficiency} impose that among clients whose loans are approved, the proportion who actually repay their loan is the same for every protected group, \textit{e.g} males and females.
    Thus, once the decision is known, the applicant's gender does not provide any information about whether they will repay the loan.
\end{example}

\begin{example}[Binary Demographic Parity: Loan Allocation]
\label{example:loan_approval2}
Referring to Example \ref{example:loan_approval1}, binary Demographic Parity requires that the ratio of loans granted ($\hat{Y} = 1$) to males is exactly equal to the ratio granted to females applicants:
\begin{equation*}
    P(\hat{Y}=1 \mid A=\text{Male}) = P(\hat{Y}=1 \mid A=\text{Female})\mathbf{.}
\end{equation*}
Because $P(\hat{Y}=0 \mid A=a) = 1 - P(\hat{Y}=1 \mid A=a)$, equalizing positive selection rates automatically guarantees equal denial rates ($P(\hat{Y}=0 \mid A=\text{Male}) = P(\hat{Y}=0 \mid A=\text{Female})$).
This ensures that applicant gender has no statistical influence on loan approval, irrespective of underlying historical repayment rates ($Y$).
\end{example}

\begin{example}[Multi-class and Multi-group Demographic Parity: Loan Allocation]
\label{example:loan_approval3}
Extending Example~\ref{example:loan_approval1} to a multi-class and multi-group setting, suppose the bank offers three distinct credit tiers: $\hat{Y} \in \{1: \text{Micro-Loan}, 2: \text{Standard Credit}, 3: \text{Premium Line}\}$, evaluated across three demographic groups $A \in \{a_1, a_2, a_3\}$.

Multi-class and Multi-group Demographic Parity mandates that the selection rate for every loan tier $\hat{y} \in \{1, 2, 3\}$ must remain identical across all demographic groups:
\begin{equation*}
    P(\hat{Y}=\hat{y} \mid A=a_1) = P(\hat{Y}=\hat{y} \mid A=a_2) = P(\hat{Y}=\hat{y} \mid A=a_3) \quad \forall \hat{y} \in \{1, 2, 3\}\mathbf{.}
\end{equation*}
If $15\%$ of applicants in group $a_1$ are granted a Premium Line ($\hat{Y}=3$), exactly $15\%$ of applicants from $a_2$ and $a_3$ must also receive a Premium Line.
This prevents demographic concentration in lower credit tiers or systemic favouritism in higher tiers.
\end{example}

\begin{example}[Binary Conditional Demographic Parity: Loan Approval]
\label{example:loan_approval4}
Revisiting Example~\ref{example:loan_approval1}, suppose credit score band $R \in \{\text{Low}, \text{Medium}, \text{High}\}$ is selected as a legitimate control variable (strata).
Binary Conditional Demographic Parity requires that within each credit score stratum $r$, loan approval rates ($\hat{Y}=1$) must be identical across protected groups:
\begin{equation*}
    P(\hat{Y}=1 \mid A=\text{Male}, R=r) = P(\hat{Y}=1 \mid A=\text{Female}, R=r) \quad \forall r \in \{\text{Low}, \text{Medium}, \text{High}\}\mathbf{.}
\end{equation*}
While overall Demographic Parity might be violated if baseline credit scores differ across groups, Conditional Demographic Parity ensures that male and female applicants with identical credit scores receive equal approval probabilities.
\end{example}

\begin{example}[Multi-class and Multi-Group Conditional Demographic Parity: Credit Allocation]
\label{example:loan_approval5}
Combining both generalizations for Example~\ref{example:loan_approval1}, consider the bank assigning applicants to three loan tiers $\hat{Y} \in \{1, 2, 3\}$ across three demographic groups $A \in \{a_1, a_2, a_3\}$, conditioned on income bracket strata $R \in \{\text{Low-Income}, \text{Middle-Income}, \text{High-Income}\}$.

Multi-class Conditional Demographic Parity requires that for every income stratum $r \in \mathcal{R}$ and every credit tier $\hat{y} \in \{1, 2, 3\}$, selection rates are equal across all demographic groups:
\begin{equation*}
\begin{aligned}
P(\hat{Y}=\hat{y} \mid A=a_1, R=r)
&= P(\hat{Y}=\hat{y} \mid A=a_2, R=r) \\
&= P(\hat{Y}=\hat{y} \mid A=a_3, R=r),
\qquad \forall \hat{y} \in \{1,2,3\}, \\
&\hspace{5.7cm} \forall r \in \mathcal{R}.
\end{aligned}
\end{equation*}
This formulation ensures that multi-tier financial products are distributed equitably across intersectional groups, controlling for legitimate economic indicators ($R$) while eliminating residual demographic bias.
\end{example}

\begin{example}[Equalized Odds \textit{vs.} Equal Accuracy: Repaying a Loan]
\label{example:eq_odds_vs_eq_acc_repayingLoan}
To illustrate this distinction, consider a toy dataset comprising four individuals: two males and two females, where exactly one male and one female are capable of repaying a loan ($Y=1$).
Requiring equal confusion matrices implies that the model must commit the exact same types of misclassification with identical probabilities for each sensitive group.
Suppose a biased classifier consistently predicts loan approval for all males ($\hat{Y}=1$) while denying all females ($\hat{Y}=0$).
The resulting confusion matrices for each demographic group are presented in Table~\ref{tab:confusion_matrices_eq_odds}.
\begin{table}[!htbp]
\begin{minipage}{0.45\textwidth}
  \centering
  \textbf{Males}
  \begin{tabularx}{\linewidth}{c >{\centering\arraybackslash}X >{\centering\arraybackslash}X}
    \toprule
     & $Y = 0$ & $Y = 1$ \\
    \midrule
    $\hat{Y} = 0$ & 0 & 0 \\
    $\hat{Y} = 1$ & 1 & 1 \\
    \bottomrule
  \end{tabularx}
\end{minipage}
\hfill
\begin{minipage}{0.45\textwidth}
  \centering
  \textbf{Females}
  \begin{tabularx}{\linewidth}{c >{\centering\arraybackslash}X >{\centering\arraybackslash}X}
    \toprule
     & $Y = 0$ & $Y = 1$ \\
    \midrule
    $\hat{Y} = 0$ & 1 & 1 \\
    $\hat{Y} = 1$ & 0 & 0 \\
    \bottomrule
  \end{tabularx}
  \centering
\end{minipage}
\caption{Group-specific confusion matrices demonstrating identical accuracy ($50\%$) despite severe error rate disparities.}
\label{tab:confusion_matrices_eq_odds}
\end{table}
Evaluating the model strictly through accuracy yields an identical score of $50\%$ for both groups, deceptively suggesting a fair predictor.
Conversely, evaluating the classifier under Equalized Odds exposes its severe discrimination against females.
\end{example}

\begin{example}[Equal Opportunity: Disease detection]
\label{example_equal_opp_disease_detection}
Disease detection is a representative application where Equal Opportunity is an appropriate fairness metric.
Consider a predictive model deployed to detect a severe medical condition, where $Y = 1$ denotes the actual presence of the disease and $\hat{Y} = 1$ indicates a positive diagnosis. 
In this high-stakes context, the primary ethical imperative is ensuring that truly sick individuals ($Y = 1$) have an equal probability of receiving a correct diagnosis ($\hat{Y} = 1$), regardless of their demographic group $A$. 

If a model fails to satisfy Equal Opportunity -- for instance, by achieving a $90\%$ True Positive Rate ($\text{TPR}$) for demographic $A$ but only $60\%$ for demographic $B$ -- it implicitly creates a severe disparity in False Negative Rates ($\text{FNR}$). 
Consequently, sick individuals in group $B$ would face a significantly higher risk ($40\%$ vs.\ $10\%$) of going undiagnosed and untreated, leading to disparate healthcare outcomes despite having identical medical needs.
\end{example}

\begin{example}[Predictive Equality: Criminal risk assessment]
\label{example:pred_equality_criminal}
Criminal risk assessment provides a classic scenario where Predictive Equality is a desirable fairness metric. 
Consider a predictive model used by courts to assess recidivism risk, where $Y = 1$ denotes that an individual will reoffend and $Y = 0$ indicates they will not, while $\hat{Y} = 1$ classifies an individual as high risk (potentially leading to bail denial or longer pre-trial detention). 
In this punitive context, a central ethical concern is avoiding disproportionate false accusations against individuals who would not reoffend ($Y = 0$).

Predictive Equality mandates that non-reoffending individuals ($Y = 0$) face an equal probability of being falsely flagged as high risk ($\hat{Y} = 1$), regardless of their demographic group $A$. 
If a model violates this criterion -- for instance, exhibiting a $10\%$ False Positive Rate ($\text{FPR}$) for group $A$ but $30\%$ for group $B$ -- it imposes a disproportionate penalty on low-risk individuals in group $B$. 
Consequently, non-reoffending individuals in group $B$ would face three times the risk of unjust detention, despite having an identical actual outcome of non-recidivism ($Y = 0$).
\end{example}

\begin{example}[Diagonal Equality: Multi-class Triage and Multi-Group Medical Assessment]
\label{example:diag_eq_medical_triage}
Consider an automated hospital triage system classifying patient urgency into three distinct classes: $Y \in \{1: \text{Routine}, 2: \text{Urgent}, 3: \text{Emergency}\}$, deployed across a multi-group demographic population $A \in \{a_1, a_2, a_3\}$. 

Diagonal Equality restricts evaluation strictly to the diagonal entries of each group's confusion matrix ($n_{kk}$).
It requires that patients who are genuinely in an \textit{Emergency} state ($Y = 3$) have an identical detection rate across all demographic groups:
\begin{equation*}
    P(\hat{Y}=3 \mid A=a_1, Y=3) = P(\hat{Y}=3 \mid A=a_2, Y=3) = P(\hat{Y}=3 \mid A=a_3, Y=3)\mathbf{.}
\end{equation*}
Simultaneously, the condition must hold independently for $Y = 2$ (\textit{Urgent}) and $Y = 1$ (\textit{Routine}). 

Suppose the classifier achieves $90\%$ recall for Class 3 among Group $a_1$, but only $60\%$ for Group $a_3$.
In that case, critically ill patients in Group $a_3$ suffer a disproportionate risk of missing immediate emergency care.
Notice, however, that this metric ignores how the remaining $40\%$ of misclassified Group $a_3$ patients are distributed across Class 1 or Class 2; it strictly demands equal true positive rates for each true label state across all demographic slices.
\end{example}

\begin{example}[Off-diagonal Equality: Multi-Tier Credit Allocation and Directional Misclassification]
\label{example:off-diag-eq}
Consider an automated credit evaluation model assigning loan applicant risk profiles into three tiers: $Y \in \{1: \text{Prime}, 2: \text{Subprime}, 3: \text{High Risk / Reject}\}$, evaluated over three intersectional demographic groups $A \in \{a_1, a_2, a_3\}$.

While Diagonal Equality regulates true positive decisions on the diagonal, Off-diagonal Equality evaluates all off-diagonal entries ($j \neq k$).
This ensures that when errors occur, their specific directional distribution does not penalize any protected group.

For instance, consider true \textit{Subprime} applicants ($Y = 2$).
A misclassification could either upgrade them to \textit{Prime} ($\hat{Y} = 1$) or downgrade them to \textit{Reject} ($\hat{Y} = 3$). 
Off-diagonal Equality requires that:
\begin{equation*}
    P(\hat{Y}=3 \mid A=a_1, Y=2) = P(\hat{Y}=3 \mid A=a_2, Y=2) = P(\hat{Y}=3 \mid A=a_3, Y=2)\mathbf{.}
\end{equation*}
If $15\%$ of true Subprime applicants in Group $a_1$ are mistakenly rejected ($\hat{Y} = 3$), but $45\%$ of Subprime applicants in Group $a_3$ are rejected, the system exhibits severe misclassification bias against Group $a_3$.
By constraining all off-diagonal cells $n_{kj}$ ($k \neq j$) to be equal across $a_1, a_2,$ and $a_3$, Off-diagonal Parity guarantees that no demographic group is systematically pushed into worse error buckets.
\end{example}

\begin{example}[Model Reliability vs.\ Sensitivity]
\label{example:reliability_vs_sensitivity}
To illustrate this distinction, consider an \gls{ml} model developed to identify diseases, where $Y=0$ denotes a healthy patient and $Y=1$ denotes a diseased patient. Suppose a classifier correctly identifies all 90 healthy patients but detects only 1 out of 10 diseased patients (Table~\ref{tab:confusion_matrix_sufficiency}).

\begin{table}[!htbp]
\centering

\renewcommand{\arraystretch}{0.85}
\begin{tabular}{lcc}
\toprule
 & \multicolumn{2}{c}{Actual outcome $Y$} \\
\cmidrule(lr){2-3}
Predicted outcome $\hat{Y}$ & Healthy ($Y=0$) & Diseased ($Y=1$) \\
\midrule
Healthy ($\hat{Y}=0$)    & 90 (TN) & 9 (FN) \\
Diseased ($\hat{Y}=1$)   & 0 (FP)  & 1 (TP) \\
\bottomrule
\end{tabular}
\caption{Confusion matrix illustrating the distinction between precision and recall: the classifier achieves perfect precision ($100\%$) but detects only $10\%$ of diseased patients.}
\label{tab:confusion_matrix_sufficiency}
\end{table}

Although the classifier performs poorly at capturing diseased patients -- achieving a True Positive Rate (recall) of only $10\%$ (equivalently, a False Negative Rate of $90\%$) -- it attains a Precision of $100\%$.
In other words, whenever the model predicts that a patient is diseased ($\hat{Y}=1$), the prediction is always correct.
Therefore, the model provides highly reliable positive predictions despite its limited sensitivity.

The previous example ignores protected groups for simplicity.
Now consider a multi-group scenario where gender is the sensitive attribute $A \in \{\text{Male}, \text{Female}\}$.
If the classifier achieves a Precision of $100\%$ for male patients but only $70\%$ for female patients, a positive prediction carries different levels of reliability depending on the patient's gender.
Such a classifier violates Sufficiency, since the predictive value of the same decision differs across demographic groups.
\end{example}

\begin{example}[Binary Predictive Parity: Automated Hiring]\
\label{example:bin_pred_parity}
Consider an automated resume screening model where $\hat{Y}=1$ indicates an invitation to an interview and $Y=1$ indicates an applicant who would actually succeed in the role.
Binary Predictive Parity demands that the \gls{ppv} be equal across protected groups.
If $80\%$ of group $a$ candidates invited to an interview ($\hat{Y}=1$) turn out to be qualified ($Y=1$), but only $50\%$ of group $b$ candidates invited are qualified, the interview invitation carries unequal predictive validity.
Recruiter confidence in positive decisions would vary by demographic group, introducing secondary bias into downstream human evaluations.
\end{example}

\begin{example}[Multi-group Multi-class Predictive Parity: Multi-Tier Credit Allocation and Directional Misclassification]
\label{example:mul_pred_parity}
Consider an automated credit evaluation model assigning loan applicant risk profiles into three tiers: $Y \in \{1: \text{Prime}, 2: \text{Subprime}, 3: \text{High Risk / Reject}\}$, evaluated over three intersectional demographic groups $A \in \{a_1, a_2, a_3\}$.

While Diagonal Equality regulates true positive decisions on the diagonal, Off-diagonal Equality evaluates all off-diagonal entries ($j \neq k$).
This ensures that when errors occur, their specific directional distribution does not penalize any protected group.

For instance, consider true \textit{Subprime} applicants ($Y = 2$).
A misclassification could either upgrade them to \textit{Prime} ($\hat{Y} = 1$) or downgrade them to \textit{Reject} ($\hat{Y} = 3$). 
Off-diagonal Equality requires that:
\begin{equation*}
    P(\hat{Y}=3 \mid A=a_1, Y=2) = P(\hat{Y}=3 \mid A=a_2, Y=2) = P(\hat{Y}=3 \mid A=a_3, Y=2)\mathbf{.}
\end{equation*}
If $15\%$ of true Subprime applicants in Group $a_1$ are mistakenly rejected ($\hat{Y} = 3$), but $45\%$ of Subprime applicants in Group $a_3$ are rejected, the system exhibits severe misclassification bias against Group $a_3$.
By constraining all off-diagonal cells $n_{kj}$ ($k \neq j$) to be equal across $a_1, a_2,$ and $a_3$, Off-diagonal Parity guarantees that no demographic group is systematically pushed into worse error buckets.
\end{example}

\begin{example}[Equal Accuracy: Masking Disparate Error Types]
\label{example:equal_acc}
Consider an automated diagnostic system evaluating medical urgency across three demographic groups $A \in \{a_1, a_2, a_3\}$ with target classes $Y \in \{1: \text{Routine}, 2: \text{Urgent}, 3: \text{Emergency}\}$.

Suppose the model achieves an identical overall accuracy of $85\%$ across all three demographic groups:
\begin{equation*}
    P(\hat{Y}=Y \mid A=a_1) = P(\hat{Y}=Y \mid A=a_2) = P(\hat{Y}=Y \mid A=a_3) = 0.85\mathbf{,}
\end{equation*}
thereby perfectly satisfying Equal Accuracy. However, examining the underlying group confusion matrices reveals stark disparities in error distribution:
for Group $a_1$, the $15\%$ total error rate consists entirely of false positives on routine cases ($\hat{Y}=2 \mid Y=1$), whereas for Group $a_3$, the $15\%$ total error rate consists entirely of dangerous false negatives on critical emergency cases ($\hat{Y}=1 \mid Y=3$).
Although Equal Accuracy reports perfect demographic parity, Group $a_3$ patients suffer life-threatening misdiagnoses, while Group $a_1$ patients face minor administrative over-referrals.
This demonstrates why Equal Accuracy is insufficient as a stand-alone fairness metric: by summing diagonal entries into a single scalar value, it obscures class-specific error distribution and directional harm across protected groups.
\end{example}

\section{Formal Proofs}\label{app:secB}
This Section provides detailed and formal proofs regarding the propositions stated in Section~\ref{sec:LABELSEC3}.

\begin{numberedproof}[Proof Equivalence to Multi-Class Equalized Odds]
\label{proof:ProofEquivalencetoMulti-ClassEqualized Odds}
Assume condition~\eqref{eq:off-diagonal-parity} holds.
\proofstep{Step 1: Total Probability Decomposition}
By the law of total probability, for any true class $j \in \mathcal{Y}$ and groups $a, b \in \mathcal{A}$, we have:
\begin{equation}
    \sum_{k \in \mathcal{Y}} P(\hat{Y} = k \mid A=a, Y = j) = 1.
\end{equation}

Isolating the correct classification term ($k = j$) from the misclassifications ($k \neq j$), we form the system:
\begin{align}
    \begin{cases}
        P(\hat{Y} = j \mid A=a, Y = j) + \sum_{k \in \mathcal{Y} \setminus \{j\}} P(\hat{Y} = k \mid A=a, Y = j) = 1 \\
        P(\hat{Y} = j \mid A=b, Y = j) + \sum_{k \in \mathcal{Y} \setminus \{j\}} P(\hat{Y} = k \mid A=b, Y = j) = 1
    \end{cases}
    \quad \forall j \in \mathcal{Y},\; \forall a, b \in \mathcal{A}.
    \label{eq:total_prob_step}
\end{align}

\proofstep{Step 2: Application of Off-diagonal Equality}
Under the Eq.~\ref{eq:off-diagonal-parity}, off-diagonal probabilities match element-wise across groups, implying that their sums are also invariant:
\begin{equation}
    \sum_{k \in \mathcal{Y} \setminus \{j\}} P(\hat{Y} = k \mid A=a, Y = j) = \sum_{k \in \mathcal{Y} \setminus \{j\}} P(\hat{Y} = k \mid A=b, Y = j).
\end{equation}

\proofstep{Step 3: Conclusion}
Subtracting the equations in \eqref{eq:total_prob_step} and simplifying via the equality above yields:
\begin{equation}
    P(\hat{Y} = j \mid A=a, Y = j) = P(\hat{Y} = j \mid A=b, Y = j) \quad \forall j \in \mathcal{Y},\; \forall a, b \in \mathcal{A}.
\end{equation}

Note that since Eq.~\ref{eq:off-diagonal-parity} enforces off-diagonal equality ($k \neq j$) and this derivation establishes diagonal equality ($k = j$), Eq.~\ref{eq:off-diagonal-parity} is mathematically equivalent to full Equalized Odds.
\end{numberedproof}

\begin{numberedproof}[Proof Mutual Exclusion of Independence and Separation]
\label{proof:MutualExclusionofIndependenceandSeparation}
\proofstep{Express the conditional prediction probability under Separation} 
Assume the Separation criterion holds ($\hat{Y} \perp A \mid Y$). By the Law of Total Probability, the scalar probability of a specific prediction $j$ given a sensitive group $a$ is:
\begin{equation}
    P(\hat{Y}=j \mid A=a) = \sum_{y \in \mathcal{Y}} P(\hat{Y}=j \mid Y=y, A=a) P(Y=y \mid A=a)
\end{equation}
Because Separation dictates that $P(\hat{Y}=j \mid Y=y, A=a) = P(\hat{Y}=j \mid Y=y)$, we can simplify the equation to:
\begin{equation}
    P(\hat{Y}=j \mid A=a) = \sum_{y \in \mathcal{Y}} P(\hat{Y}=j \mid Y=y) P(Y=y \mid A=a)
\label{eq:separation_summation}
\end{equation}

\proofstep{Transition to matrix-vector notation}
The summation in Equation~\ref{eq:separation_summation} is exactly the dot product of the $j$-th row of matrix $M$ with the column vector $\mathbf{p}_{Y \mid a}$.
Extending this to all possible predictions $j \in \hat{\mathcal{Y}}$, we can rewrite the system as a matrix-vector multiplication:
\begin{equation}
    \mathbf{p}_{\hat{Y} \mid a} = M \mathbf{p}_{Y \mid a}
\end{equation}

\proofstep{Substitute the demographic disparity vector}
From our preliminary definitions, we know that $\mathbf{p}_{Y \mid a} = \mathbf{p}_Y + \Delta \mathbf{p}_a$.
Substituting this into the matrix equation yields:
\begin{equation}
    \mathbf{p}_{\hat{Y} \mid a} = M (\mathbf{p}_Y + \Delta \mathbf{p}_a) = M \mathbf{p}_Y + M \Delta \mathbf{p}_a
		\label{eq:indep_vs_sep_matrix_eq}
\end{equation}

\proofstep{Relate the marginals}
By the Law of Total Probability without any conditional assumptions on $A$, the marginal prediction vector is $\mathbf{p}_{\hat{Y}} = M \mathbf{p}_Y$.
Substituting this back into the Equation~\ref{eq:indep_vs_sep_matrix_eq}, we obtain:
\begin{equation}
    \mathbf{p}_{\hat{Y} \mid a} = \mathbf{p}_{\hat{Y}} + M \Delta \mathbf{p}_a
\end{equation}

\proofstep{Evaluate the Independence criterion}
By hypothesis, the demographic disparity vector is not in the kernel of the matrix ($\Delta \mathbf{p}_a \notin \text{Kernel}(M)$). By definition of the kernel (null space), this implies:
\begin{equation}
    M \Delta \mathbf{p}_a \neq \mathbf{0}
\end{equation}
Consequently, inserting this inequality into the result of the previous step yields:
\begin{equation}
    \mathbf{p}_{\hat{Y} \mid a} \neq \mathbf{p}_{\hat{Y}}
\end{equation}
Translating this back to scalar probabilities, there exists at least one prediction $j$ such that $P(\hat{Y}=j \mid A=a) \neq P(\hat{Y}=j)$. This strictly violates the definition of Independence ($\hat{Y} \perp A$).
Thus, Separation and Independence cannot hold concurrently.
\end{numberedproof}

\begin{numberedproof}[Proof Coexistence of Independence and Separation]
\label{proof:CoexistenceofIndependenceandSeparation}
\proofstep{Assume Independence holds}
If Independence holds ($\hat{Y} \perp A$), the conditional probability of any prediction is equal to its marginal probability across all groups:
\begin{equation}
    P(\hat{Y}=j \mid A=a) = P(\hat{Y}=j) \quad \forall j \in \hat{\mathcal{Y}}
\end{equation}
In vector notation, this requires:
\begin{equation}
    \mathbf{p}_{\hat{Y} \mid a} = \mathbf{p}_{\hat{Y}}
\end{equation}

\proofstep{Assume Separation holds and equate the two}
As derived in Step 1 and Step 2 of Proposition \ref{prop:indep_sep_mutual_exclusion}, the assumption of Separation ($\hat{Y} \perp A \mid Y$) mathematically establishes that $\mathbf{p}_{\hat{Y} \mid a} = M \mathbf{p}_{Y \mid a}$ and $\mathbf{p}_{\hat{Y}} = M \mathbf{p}_Y$. 
Equating the expressions required for Independence, we obtain:
\begin{equation}
    M \mathbf{p}_{Y \mid a} = M \mathbf{p}_Y
\end{equation}
\begin{equation}
    M (\mathbf{p}_{Y \mid a} - \mathbf{p}_Y) = \mathbf{0}
\end{equation}
Recalling the definition of the disparity vector, this resolves to:
\begin{equation}
    M \Delta \mathbf{p}_a = \mathbf{0} \quad \forall a \in \mathcal{A}
\end{equation}
This confirms that the condition $\Delta \mathbf{p}_a \in \text{Kernel}(M)$ is the strictly necessary algebraic requirement for both criteria to coexist.

\proofstep{Derive the upper bound on the Rank of $M$}
For the condition $M \Delta \mathbf{p}_a = \mathbf{0}$ to hold with a non-zero vector $\Delta \mathbf{p}_a$ (which exists due to the premise $A \not\perp Y$), the matrix $M$ must possess a non-trivial null space.
This strictly requires $\text{Nullity}(M) \ge 1$. 
Applying the Rank-Nullity Theorem ($\text{Rank}(M) + \text{Nullity}(M) = k$), we deduce the upper bound:
\begin{equation}
    \text{Rank}(M) = k - \text{Nullity}(M) \le k - 1
		\label{eq:rank_bound_1}
\end{equation}

\proofstep{Derive the lower bound on the Rank of $M$}
We must guarantee the predictive utility premise ($\hat{Y} \not\perp Y$).
We establish this by proving that $\text{Rank}(M) = 1$ is mathematically impossible under this premise.
If $\text{Rank}(M) = 1$, all columns of the stochastic matrix $M$ must be linearly dependent.
Because $M$ is a column-stochastic matrix where every column sums to $1$, linear dependence forces all columns to be strictly identical.
Let $\mathbf{c}$ be this common column vector. This means that for any true label $y \in \mathcal{Y}$, the conditional prediction vector is exactly $\mathbf{c}$:
\begin{equation}
    \mathbf{p}_{\hat{Y} \mid y} = \mathbf{c}
\end{equation}
In scalar terms, $P(\hat{Y}=j \mid Y=y_1) = P(\hat{Y}=j \mid Y=y_2) = P(\hat{Y}=j)$ for all possible labels $y$.
This strict equality is the exact mathematical definition of statistical independence between the predictions and the true labels ($\hat{Y} \perp Y$).
Because our premise explicitly assumes $\hat{Y} \not\perp Y$, we conclude that the matrix cannot have a rank of 1.
Therefore, the lower bound must be:
\begin{equation}
    \text{Rank}(M) \ge 2
		\label{eq:rank_bound_2}
\end{equation}

\proofstep{Conclude feasibility in the Multi-class framework}
Combining the bounds \ref{eq:rank_bound_1} and \ref{eq:rank_bound_2}, we establish that the allowable rank for the classifier must fall in the interval:
\begin{equation}
    2 \le \text{Rank}(M) \le k - 1
\end{equation}
Because we are operating strictly in a multi-class setting where $k > 2$, the upper bound $k - 1$ is at least $2$.
Consequently, the interval $[2, k - 1]$ is a valid, non-empty mathematical space.
This formally proves that there exist valid matrices $M$ of rank $r \ge 2$ (ensuring predictive power, \textit{i.e.}, $\hat{Y} \not\perp Y$) that also possess a null space capable of projecting the demographic disparities $\Delta \mathbf{p}_a$ to zero.
Under these specific conditions, Independence and Separation mutually coexist.
\end{numberedproof}

\begin{numberedproof}[Proof Mutual Exclusion of Separation and Sufficiency]
\label{proof:MutualExclusionofSeparationandSufficiency}
\proofstep{Apply Bayes' Rule to the Sufficiency Condition}
By Bayes' theorem, the joint probability $P(Y=k \mid \hat{Y}=j, A=a)$ can be expressed as:
\begin{equation}
    P(Y=k \mid \hat{Y}=j, A=a) = \frac{P(\hat{Y}=j \mid Y=k, A=a) P(Y=k \mid A=a)}{P(\hat{Y}=j \mid A=a)}
\end{equation}

\proofstep{Substitute the Separation Assumption}
Assuming Separation ($\hat{Y} \perp A \mid Y$) holds, we substitute $P(\hat{Y}=j \mid Y=k, A=a) = P(\hat{Y}=j \mid Y=k)$ into the numerator:
\begin{equation}
    P(Y=k \mid \hat{Y}=j, A=a) = \frac{P(\hat{Y}=j \mid Y=k) P(Y=k \mid A=a)}{P(\hat{Y}=j \mid A=a)}
		\label{eq:separation_assumption}
\end{equation}

\proofstep{Apply the Sufficiency Assumption}
Assuming Sufficiency ($Y \perp A \mid \hat{Y}$) also holds, the conditional posterior probability must equal the marginal posterior probability $P(Y=k \mid \hat{Y}=j)$:
\begin{equation}
    P(Y=k \mid \hat{Y}=j, A=a) = P(Y=k \mid \hat{Y}=j) = \frac{P(\hat{Y}=j \mid Y=k) P(Y=k)}{P(\hat{Y}=j)}
		\label{eq:sufficiency_assumption}
\end{equation}

\proofstep{Equate Both Formulations}
Equating the Expressions \ref{eq:separation_assumption} and Expression \ref{eq:sufficiency_assumption} yields:
\begin{equation}
    \frac{P(\hat{Y}=j \mid Y=k) P(Y=k \mid A=a)}{P(\hat{Y}=j \mid A=a)} = \frac{P(\hat{Y}=j \mid Y=k) P(Y=k)}{P(\hat{Y}=j)}
\end{equation}

\proofstep{Cancel Likelihood Terms and Rearrange Ratios}
Since $P(\hat{Y}=j \mid Y=k) > 0$ by non-degeneracy, we divide both sides by $P(\hat{Y}=j \mid Y=k)$:
\begin{equation}
    \frac{P(Y=k \mid A=a)}{P(\hat{Y}=j \mid A=a)} = \frac{P(Y=k)}{P(\hat{Y}=j)}
\end{equation}
Rearranging the terms to isolate the target probabilities on the left and prediction probabilities on the right gives:
\begin{equation}
    \frac{P(Y=k \mid A=a)}{P(Y=k)} = \frac{P(\hat{Y}=j \mid A=a)}{P(\hat{Y}=j)} \quad \forall k \in \mathcal{Y}
\end{equation}

\proofstep{Deduce Independence via Ratio Constancy}
Notice that the left-hand side of the equation depends solely on $k$ and $a$, whereas the right-hand side depends solely on $j$ and $a$.
For this equality to hold for all $k \in \mathcal{Y}$, the ratio must be equal to a group-dependent constant $C_a$ that is independent of both $j$ and $k$:
\begin{equation}
    \frac{P(Y=k \mid A=a)}{P(Y=k)} = C_a \implies P(Y=k \mid A=a) = C_a P(Y=k) \quad \forall k \in \mathcal{Y}
\end{equation}
Summing both sides over all possible target classes $k \in \mathcal{Y}$:
\begin{equation}
    \sum_{k \in \mathcal{Y}} P(Y=k \mid A=a) = C_a \sum_{k \in \mathcal{Y}} P(Y=k)
\end{equation}
Because probability distributions sum to $1$ \big($\sum_k P(Y=k \mid A=a) = 1$ and $\sum_k P(Y=k) = 1$\big), we obtain $1 = C_a \cdot 1$, which proves that $C_a = 1$.

\proofstep{Conclude Contradiction}
Substituting $C_a = 1$ back into the ratio yields:
\begin{equation}
    P(Y=k \mid A=a) = P(Y=k) \quad \forall k \in \mathcal{Y}, \forall a \in \mathcal{A}
\end{equation}
This equality is the exact definition of statistical independence between the sensitive attribute and the true label ($A \perp Y$). However, this directly contradicts our foundational premise $A \not\perp Y$.

Therefore, Separation and Sufficiency cannot hold simultaneously when $A \not\perp Y$.
\end{numberedproof}

\section{Datasets Exploratory Data Analysis and Processing}\label{app:secC}
In the current literature, there exist two main types of datasets: publicly available real-world datasets and synthetic datasets. 
The former option is preferable for several key reasons.
First, real-world data provides a realistic benchmark to study model performance and fairness behaviour in natural scenarios.
Furthermore, a vast body of literature has already evaluated these datasets, establishing consensus around their adoption for comparative audits.
In contrast, purely synthetic datasets are created \textit{ad hoc} to study a desired skewed behaviour; hence, they risk being overly tailored to support a proposed thesis or novel approach.
Although synthetic datasets yield valid theoretical insights, for our experiments we selected two well-established datasets in the field of fair \gls{ml}: \textit{Adult}~\cite{adult_2} and \textit{Students' Dropout and Academic Success}~\cite{martins2021early}.

\subsection{Adult Multi-class}
\label{sec:adult_description}
\subsubsection{Overall Analysis}
Adult~\cite{adult_2}, also known as \textit{Census Income}, is a tabular dataset designed for binary classification tasks.
It is arguably the most studied benchmark dataset for \gls{ml} and, more recently, for testing fairness-aware \gls{ml} approaches.
The modelling goal is to determine whether a person, \textit{i.e.,} an instance of the dataset, earns more than $\$ 50\,000$ per year.
The dataset comprises six numerical and eight categorical features, which mainly represent demographic attributes such as age, education, relationship status, race and sex.
It gained popularity due to its size, (48\,842 total instances), and raw tabular format, which allows researchers to apply standard preprocessing workflows, including feature standardization, encoding of categorical features and data cleaning to remove or infer missing information.
Generally, scholars select \textit{Sex} as the binary sensitive attribute, which presents an imbalanced distribution: approximately $67\%$ males and $33\%$ females.
Moreover, the target distribution is skewed, with $76\%$ of the instances earning at or below $50\,000$ dollars per year, and the remaining $24\%$ earning an higher income, thus showing noticeable target class imbalance.

\subsubsection{From Binary to Multi-class}
The Census Income dataset was originally designed with a binary target.
However, for our comparative tests between native multi-class fairness and One-vs-Rest binary reduction we require a multi-class outcome.
Unfortunately, the current literature provides few real-world multi-class datasets for fairness evaluation and fewer studies regarding multi-class fairness approaches to align with.
To maintain realistic feature distributions without resorting to synthetic generation, we created a custom multi-class target for the Adult dataset by categorizing individuals into three classes based on income and education level:
\begin{equation}
    Y^{\text{custom}} = 
    \begin{cases}
        0, & \text{Income} \leq 50\text{K \$} \land \text{EducationNum} \leq 9, \\
        1, & \text{Income} \leq 50\text{K \$} \land \text{EducationNum} > 9, \\
        2, & \text{Income} > 50\text{K \$},
    \end{cases}
\end{equation}
In this formulation, $\text{Income}$ is the original binary ground target and $\text{EducationNum}$ denotes the ordinal scale of educational attainment ranging from $0$ (\textit{Preschool}) to $14$ (\textit{Doctorate}, with \textit{HS-Grad} $= 8$ and \textit{Bachelors} $= 11$).

\subsubsection{Data Processing}
First, to transform raw data into a \gls{ml}-ready format, we stripped and cleaned the categorical values from extraneous characters.
We then cleaned the dataset by dropping instances with any missing values.
As previously discussed, dropping entries may exacerbate feature or class imbalances, but modern imputation methods infer missing entries using existing attributes, which may preserve or even exacerbate underlying data biases.
Because both alternatives present inherent disadvantages, we chose the naive dropping strategy, which represents the least intrusive approach and the most widely adopted practice in other related literature works.
We strictly avoided using any fairness-enhancing preprocessing technique, because it would have modified inherent data biases and altered the result of our experiments.

To strictly prevent target leakage, both the original binary target and all education-derived attributes were discarded prior to encoding binary and multivariate categorical variables.
Because this multi-class variant was designed exclusively for tree-based ensemble models, specifically \glspl{rf}, which are inherently invariant to monotonic feature scaling, numerical standardization was unnecessary.
Subsequently, we mapped binary features into boolean values and encoded multivariate categorical features using one-hot encoding.
Furthermore, dataset splitting was not performed since the evaluation was conducted via $K$-fold \gls{cv}.

Post-cleaning, the engineered multi-class target exhibits a well-distributed support across all three categories: $39.11\%$ for Class 0, $36.09\%$ for Class 1, and $24.80\%$ for Class 2.

To evaluate our approaches, we audit three distinct sensitive attribute configurations:
\begin{itemize}
    \item \textbf{Sex:} $32.50\%$ female and $67.50\%$ male;
    \item \textbf{Race:} $86.02\%$ White and $13.98\%$ Non-White;
    \item \textbf{Intersectional (Cartesian Product):} $26.27\%$ White females, $59.75\%$ White males, $6.22\%$ Non-White females, and $7.76\%$ Non-White males.
\end{itemize}
Ultimately, we selected the binarized information whether the person was born in the United States (91.32\% born in the U.S. and 8.68\% non-U.S.) as our conditional strata feature.

\subsubsection{\textsc{FanFAIR} Evaluation}
Despite the well-established consensus surrounding the Adult dataset, we evaluated the quality level of its data using \textsc{FanFAIR}~\cite{gallese2023investigating, rispoli2025investigating, gallese2025dataset} to provide further reasoning for its adoption.
As a result of  its popularity and consensus, we set Compliance and Data Source Quality flags to the boolean value \texttt{True}.
Furthermore, we selected \texttt{IForest} as our outlier detection method and the \texttt{Hellinger-based Entropy} as our balance calculation method.

Currently, \textsc{FanFAIR}'s evaluation capability is natively restricted to binary classification, as its correlation module relies on linear association with a binary target.
To audit the multi-class dataset utilized in our experiments, we extended the \textsc{FanFAIR} framework using a One-vs-Rest transformation.
Specifically, for each class $y \in \mathcal{Y}$, we compute a standard binary audit comparing $y$ against the remaining label space $\mathcal{Y} \setminus \{y\}$.
To maintain a conservative risk bound, we aggregate these binary evaluations by taking the worst-case outcome as the final score.

Although effective, this One-vs-Rest approach artificially distorts class proportions by converting balanced multi-class distributions into skewed $1 : (|\mathcal{Y}|-1)$ binary splits, thereby penalizing dataset balance metrics, as presented in Section~\ref{sec:experimental}.
As future work, \textsc{FanFAIR} can be extended to support native multi-class auditing by replacing linear target correlations with non-parametric association metrics such as Cram\'er's V or \gls{nmi}.
These metrics quantify non-linear dependencies across arbitrary categorical label spaces without requiring binarization or inducing artificial class imbalance.

Adult Multi-class dataset reports a global $77.5\%$ score, which fits into the \texttt{good} score band, and Table~\ref{tab:fanfair_adult_multiclass} comprises the full analysis.
Despite the introduction of the custom multi-class target, the significant portion of outliers results in a score of $0\%$ for \textit{Unevenness}.
Conversely, the adoption of more equally distributed target results in an increase of the \textit{Balance} score up to $88.5\%$, which is in the Optimal score band.
While this improvement reflects a uniform underlying target distribution, it is also partly an artifact of the One-vs-Rest transformation, which evaluates binary splits ($y$ vs. $\mathcal{Y}\setminus\{y\}$) whose class proportions lie closer to a balanced $50:50$ ratio than the original binary dataset.

\begin{table}[!htbp]

\renewcommand{\arraystretch}{0.85}
\begin{tabularx}{\textwidth}{X c c}
\toprule
\textbf{Metric} & \textbf{Score} & \textbf{Score Band} \\
\midrule
Unevenness & $0.0\%$ & Critical \\
Balance & $88.5\%$ & Optimal \\
Bias with respect to Sensitive Variables & $78.5\%$ & Good \\
Compliance & $83.3\%$ & Optimal \\
Numerosity & $100.0\%$ & Optimal \\
Quality & $100.0\%$ & Optimal \\
Incompleteness & $100.0\%$ & Optimal \\
\midrule
\textbf{Overall Score} & \textbf{77.5\%} & \textbf{Good} \\
\bottomrule
\end{tabularx}
\footnotesize
\textit{Methodological setup}: Outlier detection via \texttt{IForest}; balance computed using \texttt{Hellinger-based Entropy}. \\
\textit{Sensitive Variable}: Sex, Race. \\
\textit{Compliance Parameters}: Data Protection, Copyright, Sectorial Law, Non-Discrimination, Ethics.
\centering
\caption{Adult Multi-class Dataset FanFAIR Evaluation Metrics and Compliance Overview}
\label{tab:fanfair_adult_multiclass}
\end{table}

\subsection{Students' Dropout and Academic Success}
\label{sec:students_description}
\subsubsection{Overall Analysis}
Students' Dropout and Academic Success~\cite{martins2021early} comprises administrative information from multiple academic databases into a unified dataset covering undergraduate student cohorts.
Each instance corresponds to a student and comprises $36$ numerical and categorical features covering demographic and social-economic background, enrolment details and academic performance across semesters.
The prediction task evaluates whether a student will drop out ($32.12\%$), graduate ($49.93\%$), or remain enrolled ($17.95\%$), presenting a notable class imbalance.
The dataset authors already performed a rigorous pre-processing phase, providing a ready-to-use dataset.

To evaluate our approaches, we audit three distinct sensitive attribute configurations:
\begin{itemize}
    \item \textbf{Gender:} $64.83\%$ female and $35.17\%$ male;
    \item \textbf{International Status:} $97.51\%$ non-international and $2.49\%$ international students;
    \item \textbf{Intersectional (Cartesian Product):} $63.02\%$ non-international females, $34.49\%$ non-international males, $1.81\%$ international females, and $0.68\%$ international males.
\end{itemize}
Additionally, we selected Displaced as the conditional control variable ($45.16\%$ non-displaced and $54.84\%$ displaced), a binary feature describing whether a student relocated from their permanent residence to attend university.

\subsubsection{Data Processing}
In contrast with Adult dataset, no data processing was required for Students' Dropout and Academic Success.
Consistent with the multi-class setup on the Adult dataset, model evaluation was conducted via $10$-fold \gls{cv} without reserving a static hold-out split.

\subsubsection{Data Oversampling}
To mitigate these target and subgroup representation deficits, we conducted supplementary experiments incorporating synthetic oversampling across each sensitive attribute configuration.
Specifically, we applied \gls{smotenc} strictly within each training fold to prevent data leakage, balancing all joint combinations of target classes and intersectional demographic groups.
This protocol ensures that models learn from balanced representations across all subgroup intersections while remaining evaluated on real, unmanipulated test distributions.

\subsubsection{\textsc{FanFAIR} Evaluation}
Similar to \textit{Adult} dataset, we provide the \textsc{FanFAIR} evaluation of Students' Dropout and Academic Success dataset.
To provide comparable results with \textit{Adult} dataset, we similarly select \texttt{IForest} as our outlier detection method and the \texttt{Hellinger-based Entropy} as our balance calculation method.
Moreover, we set Compliance and Data Source Quality flags to the boolean value True.

\textsc{FanFAIR} evaluation reports a global $79.4\%$ score, which fits into the \texttt{good} score band.
In Table~\ref{tab:fanfair_students} we provide results of all the metrics evaluated by the \textsc{FanFAIR} framework.
Notably, \textit{Unevenness} scores a critical $0\%$, indicating a significant presence of outliers, while demographic and target imbalances reduce the \textit{Balance} score to $76.1\%$.

\begin{table}[!htbp]

\renewcommand{\arraystretch}{0.85}
\begin{tabularx}{\textwidth}{X c c}
\toprule
\textbf{Metric} & \textbf{Score} & \textbf{Score Band} \\
\midrule
Unevenness & $0.0\%$ & Critical \\
Balance & $76.1\%$ & Good \\
Bias with respect to Sensitive Variables & $79.6\%$ & Good \\
Compliance & $83.3\%$ & Optimal \\
Numerosity & $100.0\%$ & Optimal \\
Quality & $100.0\%$ & Optimal \\
Incompleteness & $100.0\%$ & Optimal \\
\midrule
\textbf{Overall Score} & \textbf{79.4\%} & \textbf{Good} \\
\bottomrule
\end{tabularx}
\footnotesize
\textit{Methodological setup}: Outlier detection via \texttt{IForest}; balance computed using \texttt{Hellinger-based Entropy}. \\
\textit{Sensitive Variable}: Gender, International. \\
\textit{Compliance Parameters}: Data Protection, Copyright, Sectorial Law, Non-Discrimination, Ethics.
\centering
\caption{Students' Dropout and Academic Success Dataset FanFAIR Evaluation Metrics and Compliance Overview}
\label{tab:fanfair_students}
\end{table}

\section{Additional Experimental Results}
\label{app:secD}
We present a comparative empirical evaluation between native multi-class fairness metrics and standard binary reduction techniques, highlighting the analytical pitfalls and statistical distortions introduced by binarization.
To provide strong evidence of the risks and the misleading results of adopting any binarization approach, we tested both \textit{Students' Dropout and Academic Success dataset} and our multi-class \textit{Adult dataset} using three sensitive features.

\subsection{Students' Dropout and Academic Success Dataset Results}
Our first suite of experiments evaluates \textit{Students' Dropout and Academic Success} dataset, which comprise an imbalanced three-class target: Dropout ($0$), Enrolled ($1$), and Graduate ($2$).
To evaluate different demographic complexities, we test three demographic sensitive features: Gender (binary), International (binary) and their Cartesian product (intersectional multi-group).
Ultimately, Conditional \gls{dp} uses Displaced status (binary) as the stratification Control Variable.

Table~\ref{tab:t1_rf_performance} provides the standard classification performance metrics averaged across the $10-$folds for both multi-class and binary \gls{rf} models.

\begin{table}[!htbp]
\centering

\renewcommand{\arraystretch}{0.85}
\begin{tabularx}{\textwidth}{X c c c c c c}
\toprule
\textbf{Model / Target Mapping} & \textbf{Acc.} & \textbf{Bal. Acc.} & \textbf{Prec.} & \textbf{Recall} & \textbf{$F_1$} & \textbf{Spec.} \\
\midrule
Multi-class Classifier & $0.7749$ & $0.7016$ & $0.7306$ & $0.7016$ & $0.7115$ & $0.8757$ \\
Binary Classifier: Target $0$ (Dropout) & $0.8741$ & $0.8315$ & $0.8728$ & $0.7122$ & $0.7840$ & $0.9507$ \\
Binary Classifier: Target $1$ \quad (Enrolled) & $0.8271$ & $0.6673$ & $0.5223$ & $0.4180$ & $0.4639$ & $0.9165$ \\
Binary Classifier: Target $2$ (Graduate) & $0.8465$ & $0.8466$ & $0.8239$ & $0.8814$ & $0.8513$ & $0.8117$ \\
\bottomrule
\end{tabularx}
\footnotesize
\textit{Acc.}: Accuracy; 
\textit{Bal. Acc.}: Balanced Accuracy; 
\textit{Prec.}: Precision; 
\textit{Recall}: Recall; 
$F_1$: $F_1$ Score; 
\textit{Spec.}: Specificity.

\caption{\gls{rf} Classification Performance on the \textit{Students’ Dropout and Academic Success} Dataset -- Original Setup --}
\label{tab:t1_rf_performance}
\end{table}

\subsubsection{Sensitive Attribute: Gender}
Comparing the fairness metrics results in Table~\ref{tab:t1_fairness_gender} confirm that the binary reduction approaches $\text{MC}_{\text{OvR}}$ and $\text{Bin}_{\text{OvR}}$ tend to underestimate gender disparities across all target mappings.

Notably, for Target $0$ (Dropout vs. Rest), $\Delta_{\text{Pre}}$ reaches $-0.0909$ and $\Delta_{\text{Post}}$ reached$-0.0957$; meanwhile,  binary reductions present disparities around $5\%$, whereas the true multi-class disparity is a significant $14.45\%$.

Moreover, for Target $1$ (Enrolled vs. Rest), \gls{dp} and Conditional \gls{dp} highlight more prominent underestimations, with $\Delta_{\text{Pre}}$ reaching $-0.1921$ and $-0.1751$, respectively, and $\Delta_{\text{Post}}$ reaching $-0.1945$ and $-0.1843$.
This scenario showcase the intrinsic threat of any approach based on reduction to binary classification: limiting informations and combining distinct protected groups might potentially hide paramount true multi-class disparities towards these protected groups.

For Target $2$ (Graduate vs. Rest), $\Delta_{\text{Post}}$ for \gls{dp} reaches $0.0$.
This can manifest when the mapping separates the disadvantaged class from the rest, thus not altering the distribution of the most skewed protected groups which emerges unchanged as the maximum fairness discrepancy.
Although ideal, it is not possible to determine \textit{a priori} which class will have the grater disparities, and a wrong mapping might result in severe underestimation as described upon.

Ultimately, across all the three targets, multiple metrics reached gaps greater than $-0.05$. 
Although these differences may appear relatively small, they can have meaningful real-world consequences in high-stakes settings, where even modest disparities may adversely affect individuals.

\begin{table}[!htbp]
\centering

\renewcommand{\arraystretch}{0.85}
\begin{tabularx}{\textwidth}{X c c c G G}
\toprule
\textbf{Fairness Metric} & \textbf{$\text{Bin}_{\text{OvR}}$} & \textbf{$\text{MC}_{\text{OvR}}$} & \textbf{$\text{MC}_{\text{Native}}$} & \multicolumn{1}{c}{\textbf{$\Delta_{\text{Pre}}$}} & \multicolumn{1}{c}{\textbf{$\Delta_{\text{Post}}$}} \\
\midrule
\multicolumn{6}{l}{\textbf{Target 0: Dropout}} \\
\midrule
Demographic Parity & $0.1916$ & $0.1945$ & $0.2345$ & $-0.0429$ & $-0.0400$ \\
Conditional Demographic Parity & $0.2135$ & $0.2125$ & $0.2510$ & $-0.0375$ & $-0.0385$ \\
Equalized Odds & $0.0878$ & $0.0895$ & $0.1488$ & $-0.0610$ & $-0.0593$ \\
Predictive Parity & $0.0536$ & $0.0487$ & $0.1445$ & $-0.0909$ & $-0.0957$ \\
Equal Accuracy & $0.0591$ & $0.0636$ & $0.0538$ & $+0.0053$ & $+0.0098$ \\
\midrule
\multicolumn{6}{l}{\textbf{Target 1: Enrolled}} \\
\midrule
Demographic Parity & $0.0424$ & $0.0400$ & $0.2345$ & $-0.1921$ & $-0.1945$ \\
Conditional Demographic Parity & $0.0759$ & $0.0667$ & $0.2510$ & $-0.1751$ & $-0.1843$ \\
Equalized Odds & $0.0922$ & $0.0897$ & $0.1488$ & $-0.0567$ & $-0.0591$ \\
Predictive Parity & $0.0867$ & $0.1073$ & $0.1445$ & $-0.0577$ & $-0.0371$ \\
Equal Accuracy & $0.0358$ & $0.0435$ & $0.0538$ & $-0.0180$ & $-0.0103$ \\
\midrule
\multicolumn{6}{l}{\textbf{Target 2: Graduate}} \\
\midrule
Demographic Parity & $0.2292$ & $0.2345$ & $0.2345$ & $-0.0053$ & $0.0000$ \\
Conditional Demographic Parity & $0.2480$ & $0.2473$ & $0.2510$ & $-0.0030$ & $-0.0037$ \\
Equalized Odds & $0.0977$ & $0.1167$ & $0.1488$ & $-0.0512$ & $-0.0321$ \\
Predictive Parity & $0.0892$ & $0.0855$ & $0.1445$ & $-0.0553$ & $-0.0590$ \\
Equal Accuracy & $0.0208$ & $0.0312$ & $0.0538$ & $-0.0330$ & $-0.0226$ \\
\bottomrule
\end{tabularx}
\caption{Fairness Metrics Comparison for Sensitive Feature: Gender (\textit{Students’ Dropout and Academic Success} Dataset -- Original Setup --)}
\label{tab:t1_fairness_gender}
\end{table}

\subsubsection{Sensitive Attribute: International}
As shown in Table~\ref{tab:t1_fairness_international}, evaluating International student status highlights the limitations of the One-vs-Rest reductions techniques prominently, reaching significant gaps across all the mappings.
Aside from Equal Accuracy, every other metrics reach non-negligible fairness underestimations.

Notably, in Target $0$ (Dropout vs. Rest), Equalized Odds scores a gap of $-0.1914$ in $\Delta_{\text{Pre}}$ and $-0.1682$ in $\Delta_{\text{Post}}$, whereas Predictive Parity exhibits even broader discrepancies, underestimating disparity by $-0.2585$ and $-0.2292$, respectively. 

In Target $1$ (Enrolled vs. Rest), Equalized Odds reflects a negligible gap in $\Delta_{\text{Pre}}$ ($+0.0124$) but sharply deteriorates under $\Delta_{\text{Post}}$ ($-0.1525$).
Furthermore, Conditional \gls{dp} highlights consistent underestimations exceeding $-0.1952$ in that case of $\Delta_{\text{Post}}$.

Lastly, when analyzing Target $2$ (Graduate vs. Rest), Predictive Parity and Equalized Odds yield severe gaps peaking at $-0.2937$ ($\Delta_{\text{Post}}$) and $-0.2379$ ($\Delta_{\text{Pre}}$), respectively.
These findings indicate that binary reduction approaches consistently conceal substantial systemic biases affecting non-national students across all academic outcomes.

\begin{table}[!htbp]
\centering

\renewcommand{\arraystretch}{0.85}
\begin{tabularx}{\textwidth}{X c c c G G}
\toprule
\textbf{Fairness Metric} & \textbf{$\text{Bin}_{\text{OvR}}$} & \textbf{$\text{MC}_{\text{OvR}}$} & \textbf{$\text{MC}_{\text{Native}}$} & \multicolumn{1}{c}{\textbf{$\Delta_{\text{Pre}}$}} & \multicolumn{1}{c}{\textbf{$\Delta_{\text{Post}}$}} \\
\midrule
\multicolumn{6}{l}{\textbf{Target 0: Dropout}} \\
\midrule
Demographic Parity & $0.0746$ & $0.1366$ & $0.1719$ & $-0.0972$ & $-0.0352$ \\
Conditional Demographic Parity & $0.2154$ & $0.3047$ & $0.3589$ & $-0.1435$ & $-0.0542$ \\
Equalized Odds & $0.2244$ & $0.2476$ & $0.4158$ & $-0.1914$ & $-0.1682$ \\
Predictive Parity & $0.1577$ & $0.1870$ & $0.4162$ & $-0.2585$ & $-0.2292$ \\
Equal Accuracy & $0.0779$ & $0.0853$ & $0.1164$ & $-0.0385$ & $-0.0311$ \\
\midrule
\multicolumn{6}{l}{\textbf{Target 1: Enrolled}} \\
\midrule
Demographic Parity & $0.1183$ & $0.0739$ & $0.1719$ & $-0.0535$ & $-0.0980$ \\
Conditional Demographic Parity & $0.1777$ & $0.1638$ & $0.3589$ & $-0.1812$ & $-0.1952$ \\
Equalized Odds & $0.4282$ & $0.2633$ & $0.4158$ & $+0.0124$ & $-0.1525$ \\
Predictive Parity & $0.2730$ & $0.3091$ & $0.4162$ & $-0.1432$ & $-0.1071$ \\
Equal Accuracy & $0.1068$ & $0.1133$ & $0.1164$ & $-0.0096$ & $-0.0031$ \\
\midrule
\multicolumn{6}{l}{\textbf{Target 2: Graduate}} \\
\midrule
Demographic Parity & $0.1221$ & $0.1332$ & $0.1719$ & $-0.0498$ & $-0.0386$ \\
Conditional Demographic Parity & $0.2082$ & $0.2780$ & $0.3589$ & $-0.1507$ & $-0.0810$ \\
Equalized Odds & $0.1779$ & $0.1916$ & $0.4158$ & $-0.2379$ & $-0.2242$ \\
Predictive Parity & $0.1629$ & $0.1225$ & $0.4162$ & $-0.2533$ & $-0.2937$ \\
Equal Accuracy & $0.1142$ & $0.0885$ & $0.1164$ & $-0.0021$ & $-0.0279$ \\
\bottomrule
\end{tabularx}
\caption{Fairness Metrics Comparison for Sensitive Feature: International (\textit{Students’ Dropout and Academic Success} Dataset -- Original Setup --)}
\label{tab:t1_fairness_international}
\end{table}

\subsubsection{Sensitive Attribute: Gender $\times$ International}
The intersectional feature analysis in Table~\ref{tab:t1_fairness_gender_x_international} showcase the most severe misestimations.
Intersectional groups inherently suffer from representation bias, as described in Section \ref{sec:students_description}, which binary aggregation severely obscures or exacerbates.

For Target $0$ (Dropout vs. Rest), Predictive Parity produces an alarming gap of $-0.3929$ ($\Delta_{\text{Pre}}$) and $-0.2578$ ($\Delta_{\text{Post}}$), while Equalized Odds \gls{dp} peaks at $-0.2877$ ($\Delta_{\text{Pre}}$) and $-0.1826$ ($\Delta_{\text{Post}}$).

In the Target $1$ (Enrolled vs. Rest) scenario, Conditional \gls{dp} reveals a maximum gap of $-0.3929$ ($\Delta_{\text{Pre}}$) and $-0.2894$ ($\Delta_{\text{Post}}$), while Predictive Parity underperforms by up to $-0.2834$ ($\Delta_{\text{Pre}}$) and $-0.1051$ ($\Delta_{\text{Post}}$). 
This mapping is also the first scenario in which both the reduction techniques overestimate the fairness discrepancy, where Equal Accuracy picked at $0.1609$ ($\Delta_{\text{Pre}}$) and a less significant $0.0380$ ($\Delta_{\text{Post}}$).

Finally, for Target $2$ (Graduate vs. Rest), Equalized Odds underestimates the multi-class disparity by $-0.2877$ in $\Delta_{\text{Pre}}$ and $-0.1826$ in $\Delta_{\text{Post}}$, while Equal Accuracy simultaneously overestimates $\Delta_{\text{Pre}}$ by $0.0993$ and underestimates $\Delta_{\text{Post}}$ by $-0.0685$.

These results emphasize that binary reduction approaches are not structurally suited for capturing intersectional fairness, as combining subgroup predictions unpredictably inflates or deflates intersectional fairness signals.

\subsection{Students' Dropout and Academic Success with Oversampling Dataset Results}
As described in Section \ref{sec:students_description}, \textit{Students' Dropout and Academic Success} dataset suffers from severe target class and sensitive attribute representation imbalances.
To mitigate this issue we apply \gls{smotenc} to rebalance the intersection of the target class and protected features.
The empirical evaluations below show that oversampling alone fails to resolve systematic underestimation; therefore, in order to overcome these unfairness, more advanced pre-processing fairness-enhancing methodologies are needed.

We report the experimental evaluation conducted across identical sensitive features, conditional attribute, and target binarization schemes to provide a complete and robust comparison.
Table~\ref{tab:t2_rf_performance} provides the predictive performance of \glspl{rf} averaged over $10-$fold evaluations.
\begin{table}[!htbp]
\centering

\renewcommand{\arraystretch}{0.85}
\begin{tabularx}{\textwidth}{X c c c c c c}
\toprule
\textbf{Model / Target Mapping} & \textbf{Acc.} & \textbf{Bal. Acc.} & \textbf{Prec.} & \textbf{Recall} & \textbf{$F_1$} & \textbf{Spec.} \\
\midrule
Multi-class Classifier & $0.7502$ & $0.7038$ & $0.7055$ & $0.7038$ & $0.7020$ & $0.8729$ \\
Binary Classifier: Target $0$ (Dropout) & $0.8707$ & $0.8345$ & $0.8446$ & $0.7333$ & $0.7846$ & $0.9357$ \\
Binary Classifier: Target $1$ \quad (Enrolled) & $0.7963$ & $0.7018$ & $0.4459$ & $0.5543$ & $0.4938$ & $0.8493$ \\
Binary Classifier: Target $2$ (Graduate) & $0.8388$ & $0.8388$ & $0.8381$ & $0.8393$ & $0.8386$ & $0.8384$ \\
\bottomrule
\end{tabularx}
\footnotesize
\textit{Acc.}: Accuracy; 
\textit{Bal. Acc.}: Balanced Accuracy; 
\textit{Prec.}: Precision; 
\textit{Recall}: Recall; 
$F_1$: $F_1$ Score; 
\textit{Spec.}: Specificity.

\caption{\gls{rf} Classification Performance on the \textit{Students’ Dropout and Academic Success} Dataset -- Oversampled Setup --}
\label{tab:t2_rf_performance}
\end{table}

\subsubsection{Sensitive Attribute: Gender}
As presented in Table~\ref{tab:t2_fairness_gender}, when evaluating Target $0$ (Dropout vs. Rest), Predictive Parity exhibits persistent negative gaps exceeding $-0.09$, whereas Equalized Odds underestimations cluster around $-0.07$.

In the Target $1$ (Enrolled vs. Rest) scenario, \gls{dp} and Conditional \gls{dp} remain significantly distorted, recording gaps of $-0.1895$ ($\Delta_{\text{Pre}}$), $-0.1812$ ($\Delta_{\text{Post}}$) for \gls{dp}, and $-0.1733$ ($\Delta_{\text{Pre}}$) and $-0.1622$ ($\Delta_{\text{Post}}$) for Conditional \gls{dp}.

For Target $2$ (Graduate vs. Rest) configuration, while most disparity gaps narrow down, Equalized Odds persistently underestimates bias with gaps near $-0.07$.
\begin{table}[!htbp]
\centering

\renewcommand{\arraystretch}{0.85}
\begin{tabularx}{\textwidth}{X c c c G G}
\toprule
\textbf{Fairness Metric} & \textbf{$\text{Bin}_{\text{OvR}}$} & \textbf{$\text{MC}_{\text{OvR}}$} & \textbf{$\text{MC}_{\text{Native}}$} & \multicolumn{1}{c}{\textbf{$\Delta_{\text{Pre}}$}} & \multicolumn{1}{c}{\textbf{$\Delta_{\text{Post}}$}} \\
\midrule
\multicolumn{6}{l}{\textbf{Target 0: Dropout}} \\
\midrule
Demographic Parity & $0.1741$ & $0.1850$ & $0.2147$ & $-0.0406$ & $-0.0297$ \\
Conditional Demographic Parity & $0.1968$ & $0.1988$ & $0.2380$ & $-0.0412$ & $-0.0392$ \\
Equalized Odds & $0.0939$ & $0.0983$ & $0.1651$ & $-0.0712$ & $-0.0668$ \\
Predictive Parity & $0.0387$ & $0.0369$ & $0.1326$ & $-0.0939$ & $-0.0957$ \\
Equal Accuracy & $0.0698$ & $0.0678$ & $0.0610$ & $+0.0087$ & $+0.0068$ \\
\midrule
\multicolumn{6}{l}{\textbf{Target 1: Enrolled}} \\
\midrule
Demographic Parity & $0.0252$ & $0.0335$ & $0.2147$ & $-0.1895$ & $-0.1812$ \\
Conditional Demographic Parity & $0.0646$ & $0.0758$ & $0.2380$ & $-0.1733$ & $-0.1622$ \\
Equalized Odds & $0.0965$ & $0.1086$ & $0.1651$ & $-0.0686$ & $-0.0565$ \\
Predictive Parity & $0.0885$ & $0.0787$ & $0.1326$ & $-0.0441$ & $-0.0539$ \\
Equal Accuracy & $0.0501$ & $0.0442$ & $0.0610$ & $-0.0109$ & $-0.0169$ \\
\midrule
\multicolumn{6}{l}{\textbf{Target 2: Graduate}} \\
\midrule
Demographic Parity & $0.2086$ & $0.2109$ & $0.2147$ & $-0.0061$ & $-0.0038$ \\
Conditional Demographic Parity & $0.2247$ & $0.2355$ & $0.2380$ & $-0.0133$ & $-0.0025$ \\
Equalized Odds & $0.0951$ & $0.0935$ & $0.1651$ & $-0.0699$ & $-0.0715$ \\
Predictive Parity & $0.0972$ & $0.0940$ & $0.1326$ & $-0.0354$ & $-0.0386$ \\
Equal Accuracy & $0.0331$ & $0.0269$ & $0.0610$ & $-0.0279$ & $-0.0341$ \\
\bottomrule
\end{tabularx}
\caption{Fairness Metrics Comparison for Sensitive Feature: Gender (\textit{Students’ Dropout and Academic Success} Dataset -- Oversampled Setup --)}
\label{tab:t2_fairness_gender}
\end{table}

\subsubsection{Sensitive Attribute: International}
Table~\ref{tab:t2_fairness_international} presents disparity gaps under oversampled regime for the \textit{International} sensitive feature.

For Target $0$ (Dropout vs. Rest), the Predictive Parity disparities gaps remain high at $-0.2081$ for $\Delta_{\text{Pre}}$ and $-0.1959$ for $\Delta_{\text{Post}}$, whereas Equalized Odds produces underestimation gaps exceeding $-0.11$.

In the Target $1$ (Enrolled vs. Rest) scenario, Equalized Odds reaches underestimation gaps of $-0.1832$ ($\Delta_{\text{Pre}}$) and $-0.1775$ ($\Delta_{\text{Post}}$), while Conditional \gls{dp} gaps widen to approximately $-0.13$.

Lastly, the Target $2$ (Graduate vs. Rest) configuration yields the most alarming gaps, with Predictive Parity reaching $-0.2770$ ($\Delta_{\text{Pre}}$) and $-0.2670$ ($\Delta_{\text{Post}}$).

\begin{table}[!htbp]
\centering

\renewcommand{\arraystretch}{0.85}
\begin{tabularx}{\textwidth}{X c c c G G}
\toprule
\textbf{Fairness Metric} & \textbf{$\text{Bin}_{\text{OvR}}$} & \textbf{$\text{MC}_{\text{OvR}}$} & \textbf{$\text{MC}_{\text{Native}}$} & \multicolumn{1}{c}{\textbf{$\Delta_{\text{Pre}}$}} & \multicolumn{1}{c}{\textbf{$\Delta_{\text{Post}}$}} \\
\midrule
\multicolumn{6}{l}{\textbf{Target 0: Dropout}} \\
\midrule
Demographic Parity & $0.0944$ & $0.1120$ & $0.1564$ & $-0.0619$ & $-0.0444$ \\
Conditional Demographic Parity & $0.2194$ & $0.2362$ & $0.2907$ & $-0.0714$ & $-0.0545$ \\
Equalized Odds & $0.2297$ & $0.2490$ & $0.3609$ & $-0.1312$ & $-0.1118$ \\
Predictive Parity & $0.1964$ & $0.2086$ & $0.4046$ & $-0.2081$ & $-0.1959$ \\
Equal Accuracy & $0.0719$ & $0.0830$ & $0.0941$ & $-0.0223$ & $-0.0111$ \\
\midrule
\multicolumn{6}{l}{\textbf{Target 1: Enrolled}} \\
\midrule
Demographic Parity & $0.0739$ & $0.0648$ & $0.1564$ & $-0.0825$ & $-0.0915$ \\
Conditional Demographic Parity & $0.1618$ & $0.1529$ & $0.2907$ & $-0.1289$ & $-0.1378$ \\
Equalized Odds & $0.1777$ & $0.1833$ & $0.3609$ & $-0.1832$ & $-0.1775$ \\
Predictive Parity & $0.3667$ & $0.3474$ & $0.4046$ & $-0.0378$ & $-0.0572$ \\
Equal Accuracy & $0.0743$ & $0.0822$ & $0.0941$ & $-0.0198$ & $-0.0119$ \\
\midrule
\multicolumn{6}{l}{\textbf{Target 2: Graduate}} \\
\midrule
Demographic Parity & $0.1314$ & $0.1359$ & $0.1564$ & $-0.0249$ & $-0.0205$ \\
Conditional Demographic Parity & $0.2443$ & $0.2617$ & $0.2907$ & $-0.0464$ & $-0.0290$ \\
Equalized Odds & $0.2180$ & $0.2348$ & $0.3609$ & $-0.1428$ & $-0.1260$ \\
Predictive Parity & $0.1276$ & $0.1376$ & $0.4046$ & $-0.2770$ & $-0.2670$ \\
Equal Accuracy & $0.0922$ & $0.0960$ & $0.0941$ & $-0.0020$ & $+0.0019$ \\
\bottomrule
\end{tabularx}
\caption{Fairness Metrics Comparison for Sensitive Feature: International (\textit{Students’ Dropout and Academic Success} Dataset -- Oversampled Setup --)}
\label{tab:t2_fairness_international}
\end{table}

\subsubsection{Sensitive Attribute: Gender $\times$ International}
These trends, supported by Table \ref{tab:t2_fairness_gender_x_international}, confirm that rebalancing class sample sizes through oversampling does not compensate for the loss of information caused by collapsing of multi-class targets into binary targets.

In Target $0$ (Dropout vs. Rest), Equalized Odds exhibits the  largest underestimation gaps reaching $-0.4247$ in $\Delta_{\text{Pre}}$ and $-0.3970$ in $\Delta_{\text{Post}}$.
This highlights a critical information loss, given that native multi-class baseline ($\text{MC}_{\text{Native}}$) scores an alarming true multi-class fairness disparity of $+0.9003$.
Notably, Equal Accuracy scores a positive overestimation $\Delta_{\text{Pre}}$ gap of $+0.0444$.

Similarly, in Target $1$ (Enrolled vs. Rest) configuration, the fairness gaps results non-negligible across all evaluated fairness metrics.
\gls{dp} yields gaps of $-0.1837$ ($\Delta_{\text{Pre}}$) and $-0.1513$ ($\Delta_{\text{Post}}$), whereas Conditional \gls{dp} gaps expand significantly to alarming $-0.4438$ and $-0.2971$, respectively.
Moreover, Equalized Odds consistently peaks disparity gaps over $-0.20$ in underestimation, while Predictive Parity achieves $-0.0984$ ($\Delta_{\text{Pre}}$) and $-0.2384$ ($\Delta_{\text{Post}}$).
Consistent with  Target $0$ (Dropout vs. Rest) settings, Equal Accuracy results in a positive $\Delta_{\text{Pre}}$ gap of $+0.0525$.

Likewise, for Target $2$ (Graduate vs. Rest) scenario, Equalized Odds gaps remains severe at $-0.4010$ ($\Delta_{\text{Pre}}$) $-0.3265$ ($\Delta_{\text{Post}}$), hiding the true multi-class fairness disparities of $0.9003$ ($\text{MC}_{\text{Native}}$).
This is followed by Predictive Parity, which reports underestimation gaps of $-0.2948$ in $\Delta_{\text{Pre}}$ and of $-0.2375$ $\Delta_{\text{Post}}$.
The remaining metrics similarly exhibits non-negligible gaps, with Equal Accuracy scoring $\Delta_{\text{Post}}$ overestimations equal to $+0.0687$.
\begin{table}[!htbp]
\centering

\renewcommand{\arraystretch}{0.85}
\begin{tabularx}{\textwidth}{X c c c G G}
\toprule
\textbf{Fairness Metric} & \textbf{$\text{Bin}_{\text{OvR}}$} & \textbf{$\text{MC}_{\text{OvR}}$} & \textbf{$\text{MC}_{\text{Native}}$} & \multicolumn{1}{c}{\textbf{$\Delta_{\text{Pre}}$}} & \multicolumn{1}{c}{\textbf{$\Delta_{\text{Post}}$}} \\
\midrule
\multicolumn{6}{l}{\textbf{Target 0: Dropout}} \\
\midrule
Demographic Parity & 0.2571 & 0.3458 & 0.4359 & -0.1787 & -0.0901 \\
Conditional Demographic Parity & 0.5536 & 0.6689 & 0.7836 & -0.2300 & -0.1147 \\
Equalized Odds & 0.4756 & 0.5033 & 0.9003 & -0.4247 & -0.3970 \\
Predictive Parity & 0.4954 & 0.3934 & 0.6340 & -0.1386 & -0.2406 \\
Equal Accuracy & 0.2708 & 0.2385 & 0.2264 & 0.0444 & 0.0121 \\
\midrule
\multicolumn{6}{l}{\textbf{Target 1: Enrolled}} \\
\midrule
Demographic Parity & 0.2521 & 0.2846 & 0.4359 & -0.1837 & -0.1513 \\
Conditional Demographic Parity & 0.3398 & 0.4865 & 0.7836 & -0.4438 & -0.2971 \\
Equalized Odds & 0.6941 & 0.6646 & \textbf{0.9003} & -0.2062 & -0.2357 \\
Predictive Parity & 0.5356 & 0.3956 & 0.6340 & -0.0984 & -0.2384 \\
Equal Accuracy & 0.2789 & 0.2240 & 0.2264 & 0.0525 & -0.0024 \\
\midrule
\multicolumn{6}{l}{\textbf{Target 2: Graduate}} \\
\midrule
Demographic Parity & 0.3191 & 0.3579 & 0.4359 & -0.1167 & -0.0780 \\
Conditional Demographic Parity & 0.6178 & 0.6613 & 0.7836 & -0.1658 & -0.1223 \\
Equalized Odds & 0.4993 & 0.5738 & \textbf{0.9003} & -0.4010 & -0.3265 \\
Predictive Parity & 0.3392 & 0.3965 & 0.6340 & -0.2948 & -0.2375 \\
Equal Accuracy & 0.2502 & 0.2950 & 0.2264 & 0.0238 & 0.0687 \\
\bottomrule
\end{tabularx}
\footnotesize
\textit{Acc.}: Accuracy; 
\textit{Bal. Acc.}: Balanced Accuracy; 
\textit{Prec.}: Precision; 
\textit{Recall}: Recall; 
$F_1$: $F_1$ Score; 
\textit{Spec.}: Specificity.
\caption{Fairness Metrics Comparison for Sensitive Feature: Gender $\times$ International (\textit{Students’ Dropout and Academic Success} Dataset -- Oversampled Setup --)}
\label{tab:t2_fairness_gender_x_international}
\end{table}

\subsection{Adult Multi-class Dataset Results}
As described in Section \ref{sec:adult_description}, the Adult dataset is originally designed for binary classification.
We extend it to a the multi-class classification tasks by combining the original target and the feature \textit{EducationNum}, thus obtaining the following three balanced classes:  \textit{Low-Income and Low-Education}, defined as $Income \leq 50\text{K \$} \land Education Num \leq 9$ (Target 0); \textit{Low-Income and High-Education}, defined as $Income \leq 50\text{K \$} \land Education Num > 9$ (Target 1); and \textit{High-Income}, simply defined as $Income > 50\text{K \$}$ (Target 2).

Similarly to previous dataset, we choose three sensitive attribute to provide detailed and robust experiments: \textit{Sex} (binary), \textit{Race -- White} (Binary) and their intersection (multi-group).
Lastly, we select whether an individual was born in the United States (\textit{Country\_United\_States}, binary as well) as our conditional attribute.

Table~\ref{tab:t3_rf_performance} shows the performances of the \glspl{rf} across all target configurations.
\begin{table}[!htbp]
\centering

\renewcommand{\arraystretch}{0.85}
\begin{tabularx}{\textwidth}{X c c c c c c}
\toprule
\textbf{Model / Target Mapping} & \textbf{Acc.} & \textbf{Bal. Acc.} & \textbf{Prec.} & \textbf{Recall} & \textbf{$F_1$} & \textbf{Spec.} \\
\midrule
Multi-class Classifier & $0.6559$ & $0.6662$ & $0.6553$ & $0.6662$ & $0.6572$ & $0.8251$ \\
Binary Classifier: Target $0$ (Low-Income \& Low-Education) & $0.7324$ & $0.7392$ & $0.6290$ & $0.7703$ & $0.6925$ & $0.7081$ \\
Binary Classifier: Target $1$ (Low-Income \& High-Education) & $0.7263$ & $0.7018$ & $0.6226$ & $0.6138$ & $0.6181$ & $0.7899$ \\
Binary Classifier: Target $2$ (High-Income) & $0.8391$ & $0.8096$ & $0.6526$ & $0.7510$ & $0.6983$ & $0.8682$ \\
\bottomrule
\end{tabularx}
\footnotesize
\textit{Acc.}: Accuracy; 
\textit{Bal. Acc.}: Balanced Accuracy; 
\textit{Prec.}: Precision; 
\textit{Recall}: Recall; 
$F_1$: $F_1$ Score; 
\textit{Spec.}: Specificity.

\caption{\gls{rf} Classification Performance on the \textit{Adult Multi-class} Dataset}
\label{tab:t3_rf_performance}
\end{table}

\subsubsection{Sensitive Attribute: Sex}
The fairness evaluation with \textit{Sex} as the protected group (Table~\ref{tab:t3_fairness_sex}) underscores how target mapping selection drastically dictates the observed disparity gaps.

First, in Target $0$ (Low-Income \& Low-Education vs. Rest) scenario, the binary \gls{rf} classifier reports a relatively low fairness discrepancy ($\text{Bin}_{\text{OvR}}$), whereas the native multi-class model reveals  severe discrepancy ($\text{MC}_{\text{Native}}$).
Notably, \gls{dp} suffers from significant underestimation, yielding gaps of $-0.2993$ and $-0.2741$ for $\Delta_{\text{Pre}}$ and $\Delta_{\text{Post}}$, respectively.
Similarly, Conditional \gls{dp} yields underestimations of $-0.2857$ ($\Delta_{\text{Pre}}$) and $-0.2551$ ($\Delta_{\text{Post}}$), whereas Equalized Odds gaps raise to $-0.2646$ ($\Delta_{\text{Pre}}$) and $-0.2428$ ($\Delta_{\text{Post}}$).

Conversely, in Target $1$ (Low-Income \& Low-Education vs. Rest) configuration, \gls{dp} and Conditional \gls{dp} both show minimal positive $\Delta_{\text{Pre}}$ gaps ($+0.0221$ and $+0.0211$, respectively), whereas the remaining disparity gaps results fall below $+0.02$ and thus negligible.
This happens because this specific binary mapping isolates the subgroups where disparity skew the most, leaving remaining joined groups negligibly altered.

For Target $2$ (High-Income vs. Rest), the results showcase significant underestimations where Equalized Odds scores gaps of $-0.1778$ ($\Delta_{\text{Pre}}$) and $-0.1564$ ($\Delta_{\text{Post}}$).
\gls{dp} and Conditional \gls{dp} both exhibit gaps of approximately $-0.061$ $\Delta_{\text{Pre}}$ and $-0.048$ in $\Delta_{\text{Post}}$, whereas Equal Accuracy overestimates disparity yielding positive gaps of $+0.0872$ ($\Delta_{\text{Pre}}$) and $+0.0881$($\Delta_{\text{Post}}$).

These results effectively showcase that selecting an arbitrary binary reduction can severely mask true disparities or, conversely, induce false overestimations.

\subsubsection{Sensitive Attribute: Race -- White}
Table~\ref{tab:t3_fairness_race_white} presents the fairness evaluations for the Race -- White sensitive attribute (White vs. Non-White).

In the Target $0$ (Low-Income \& Low-Education vs. Rest) scenario, the \gls{dp} and Equalized Odds $\Delta_{\text{Pre}}$ gaps are the only pronounced distortions, recording $-0.0625$ and $-0.0559$, respectively.
The remaining gaps scores lower than $-0.05$ an therefore can be considered relatively negligible.

In Target $1$ (Low-Income \& Low-Education vs. Rest) settings, \gls{dp} gaps widen beyond $-0.1$ with both $\text{Bin}_{\text{OvR}}$ and $\text{MC}_{\text{OvR}}$, close to achieving fair scores; similarly, Conditional \gls{dp} reaches $-0.0819$ ($\Delta_{\text{Pre}}$) and $-0.0895$ ($\Delta_{\text{Post}}$).
Lastly, Equalized Odds records $-0.0816$ $\Delta_{\text{Pre}}$ and $-0.0589$ $\Delta_{\text{Post}}$.

The Target $2$ (High-Income vs. Rest) mapping narrows down all the gap absolute values below $0.05$, with Equal Accuracy remaining the only metric with positive overestimations.

Although less alarming, this setting reinforces once more the inherent risks of arbitrary target binarization, which can obscure substantial multi-class bias as observed in Target $1$ (Low-Income \& Low-Education vs. Rest).

\begin{table}[!htbp]
\centering

\renewcommand{\arraystretch}{0.85}
\begin{tabularx}{\textwidth}{X c c c G G}
\toprule
\textbf{Fairness Metric} & \textbf{$\text{Bin}_{\text{OvR}}$} & \textbf{$\text{MC}_{\text{OvR}}$} & \textbf{$\text{MC}_{\text{Native}}$} & \multicolumn{1}{c}{\textbf{$\Delta_{\text{Pre}}$}} & \multicolumn{1}{c}{\textbf{$\Delta_{\text{Post}}$}} \\
\midrule
\multicolumn{6}{l}{\textbf{Target 0: Low-Income \& Low-Education}} \\
\midrule
Demographic Parity & $0.0831$ & $0.1024$ & $0.1455$ & $-0.0625$ & $-0.0431$ \\
Conditional Demographic Parity & $0.1975$ & $0.2078$ & $0.2112$ & $-0.0138$ & $-0.0035$ \\
Equalized Odds & $0.0633$ & $0.0862$ & $0.1192$ & $-0.0559$ & $-0.0330$ \\
Predictive Parity & $0.0655$ & $0.0487$ & $0.0671$ & $-0.0016$ & $-0.0184$ \\
Equal Accuracy & $0.0167$ & $0.0147$ & $0.0309$ & $-0.0143$ & $-0.0162$ \\
\midrule
\multicolumn{6}{l}{\textbf{Target 1: Low-Income \& High-Education}} \\
\midrule
Demographic Parity & $0.0296$ & $0.0431$ & $0.1455$ & $-0.1160$ & $-0.1024$ \\
Conditional Demographic Parity & $0.1293$ & $0.1217$ & $0.2112$ & $-0.0819$ & $-0.0895$ \\
Equalized Odds & $0.0376$ & $0.0603$ & $0.1192$ & $-0.0816$ & $-0.0589$ \\
Predictive Parity & $0.0378$ & $0.0338$ & $0.0671$ & $-0.0293$ & $-0.0333$ \\
Equal Accuracy & $0.0121$ & $0.0153$ & $0.0309$ & $-0.0188$ & $-0.0156$ \\
\midrule
\multicolumn{6}{l}{\textbf{Target 2: High-Income}} \\
\midrule
Demographic Parity & $0.1394$ & $0.1455$ & $0.1455$ & $-0.0061$ & $0.0000$ \\
Conditional Demographic Parity & $0.1704$ & $0.1766$ & $0.2112$ & $-0.0408$ & $-0.0346$ \\
Equalized Odds & $0.0947$ & $0.1016$ & $0.1192$ & $-0.0245$ & $-0.0177$ \\
Predictive Parity & $0.0301$ & $0.0325$ & $0.0671$ & $-0.0370$ & $-0.0346$ \\
Equal Accuracy & $0.0614$ & $0.0636$ & $0.0309$ & $+0.0305$ & $+0.0327$ \\
\bottomrule
\end{tabularx}
\caption{Fairness Metrics Comparison for Sensitive Feature: Race -- White (\textit{Adult Multi-class} Dataset)}
\label{tab:t3_fairness_race_white}
\end{table}

\subsubsection{Sensitive Attribute: Sex $\times$ Race -- White}
Finally, analyzing the combined Sex $\times$ Race -- White sensitive attributes (Table~\ref{tab:t3_fairness_sex_x_race_white}) reinforces the trend observed in the previous intersectional experiments.

Target $0$ (Low-Income \& Low-Education vs. Rest) showcase the worst scenario: \gls{dp} gaps reach $-0.2206$ ($\Delta_{\text{Pre}}$) and $-0.1724$ ($\Delta_{\text{Post}}$), and similarly Equalized Odds yields underestimations of $0.2112$ and $-0.1626$, respectively.

For Target $1$ (Low-Income \& Low-Education vs. Rest), all gap reductions are negligible, with the sole exception of Predictive Parity, which exhibits $-0.0568$ in $\Delta_{\text{Pre}}$ and $-0.0640$ $\Delta_{\text{Post}}$.

Finally, in Target $2$ (High-Income vs. Rest), equalized Odds yields gaps of $-0.1023$ ($\Delta_{\text{Pre}}$) and $-0.0891$ ($\Delta_{\text{Post}}$), whereas Equal Accuracy significantly overestimates disparities, reaching $+0.0833$ and $+0.0860$, respectively.

\begin{table}[!htbp]
\centering

\renewcommand{\arraystretch}{0.85}
\begin{tabularx}{\textwidth}{X c c c G G}
\toprule
\textbf{Fairness Metric} & \textbf{$\text{Bin}_{\text{OvR}}$} & \textbf{BAM Mean} & \textbf{$\text{MC}_{\text{Native}}$} & \multicolumn{1}{c}{\textbf{$\Delta_{\text{Pre}}$}} & \multicolumn{1}{c}{\textbf{Gap $\Delta_{\text{Post}}$}} \\
\midrule
\multicolumn{6}{l}{\textbf{Target 0: Low-Income \& Low-Education}} \\
\midrule
Demographic Parity & $0.1256$ & $0.1738$ & $0.3462$ & $-0.2206$ & $-0.1724$ \\
Conditional Demographic Parity & $0.2591$ & $0.2634$ & $0.3768$ & $-0.1177$ & $-0.1134$ \\
Equalized Odds & $0.1498$ & $0.1984$ & $0.3610$ & $-0.2112$ & $-0.1626$ \\
Predictive Parity & $0.0977$ & $0.0927$ & $0.1596$ & $-0.0619$ & $-0.0669$ \\
Equal Accuracy & $0.0482$ & $0.0407$ & $0.0643$ & $-0.0161$ & $-0.0236$ \\
\midrule
\multicolumn{6}{l}{\textbf{Target 1: Low-Income \& High-Education}} \\
\midrule
Demographic Parity & $0.3647$ & $0.3390$ & $0.3462$ & $+0.0185$ & $-0.0072$ \\
Conditional Demographic Parity & $0.3894$ & $0.3689$ & $0.3768$ & $+0.0126$ & $-0.0079$ \\
Equalized Odds & $0.3555$ & $0.3575$ & $0.3610$ & $-0.0055$ & $-0.0035$ \\
Predictive Parity & $0.1028$ & $0.0956$ & $0.1596$ & $-0.0568$ & $-0.0640$ \\
Equal Accuracy & $0.0462$ & $0.0577$ & $0.0643$ & $-0.0181$ & $-0.0067$ \\
\midrule
\multicolumn{6}{l}{\textbf{Target 2: High-Income}} \\
\midrule
Demographic Parity & $0.3204$ & $0.3313$ & $0.3462$ & $-0.0258$ & $-0.0150$ \\
Conditional Demographic Parity & $0.3408$ & $0.3526$ & $0.3768$ & $-0.0360$ & $-0.0242$ \\
Equalized Odds & $0.2587$ & $0.2719$ & $0.3610$ & $-0.1023$ & $-0.0891$ \\
Predictive Parity & $0.1266$ & $0.1338$ & $0.1596$ & $-0.0330$ & $-0.0258$ \\
Equal Accuracy & $0.1476$ & $0.1503$ & $0.0643$ & $+0.0833$ & $+0.0860$ \\
\bottomrule
\end{tabularx}
\caption{Fairness Metrics Comparison for Sensitive Feature: Sex $\times$ Race (White) (Adult Multi-class)}
\label{tab:t3_fairness_sex_x_race_white}
\end{table}

\end{document}